\documentclass[lettersize,journal]{IEEEtran}
\usepackage{amsmath,amsfonts}
\usepackage{algorithmic}
\usepackage{algorithm}
\usepackage{array}
\usepackage{ctable}
\usepackage[caption=false,font=normalsize,labelfont=sf,textfont=sf]{subfig}
\usepackage{textcomp}
\usepackage{stfloats}
\usepackage{url}
\usepackage{graphicx}
\usepackage{colortbl}
\usepackage{xcolor}
\usepackage{vcell}
\usepackage{hyperref}
\usepackage{url}
\usepackage{multirow}
\usepackage[numbers]{natbib}

\usepackage{booktabs}   

\DeclareMathOperator{\sgn}{sgn}
\newcommand*{\myalign}[2]{\multicolumn{1}{#1}{#2}}

\begin{document}

\title{Deep Learning and Symbolic Regression for Discovering Parametric Equations}

\author{Michael~Zhang$^{*1}$, Samuel~Kim$^{*1,3,a}$, Peter~Y.~Lu$^{2,4}$, Marin~Solja\v{c}i\'{c}$^{2,b}$
\thanks{$^*$These authors contributed equally to this work.}
\thanks{$^1$Department of Electrical Engineering and Computer Science, Massachusetts Institute of Technology, Cambridge, MA, USA}
\thanks{$^2$Department of Physics, Massachusetts Institute of Technology, Cambridge, MA, USA}
\thanks{$^3$Currently at Department of Research and Exploratory Development, Johns Hopkins University Applied Physics Laboratory, Laurel, MD, USA}%
\thanks{$^3$Currently at Data Science Institute, University of Chicago, Chicago, IL, USA}
\thanks{$^a$E-mail: samkim@mit.edu}
\thanks{$^b$E-mail: soljacic@mit.edu}
}



\maketitle

\begin{abstract}
Symbolic regression is a machine learning technique that can learn the governing formulas of data and thus has the potential to transform scientific discovery. However, symbolic regression is still limited in the complexity and dimensionality of the systems that it can analyze. Deep learning on the other hand has transformed machine learning in its ability to analyze extremely complex and high-dimensional datasets. We propose a neural network architecture to extend symbolic regression to parametric systems where some coefficient may vary but the structure of the underlying governing equation remains constant. We demonstrate our method on various analytic expressions and partial differential equations (PDEs) with varying coefficients, and show that it extrapolates well outside of the training domain. The proposed neural network-based architecture can also be enhanced by integrating with other deep learning architectures such that it can analyze high-dimensional data while being trained end-to-end. To this end, we demonstrate the scalability of our architecture by incorporating a convolutional encoder to analyze 1D images of varying spring systems.
\end{abstract}


\section{Introduction}
\IEEEPARstart{C}{omplex} systems can often be described by relatively simple and interpretable mathematical equations, ranging from Maxwell's equations for electrodynamics  \cite{Griffiths:1492149} to Hooke's law for harmonic oscillators.
Thus, the discovery of the governing equations for natural and artificial systems from data is key to many scientific and engineering disciplines.
While scientists and engineers have often spent years developing insights to discover such equations, machine learning has become alluring in its potential to tackle and automate extremely complex tasks.
In particular, symbolic regression is a machine learning technique that searches for mathematical expressions that best fit the data, ideally resulting in a model that is interpretable and explains the underlying dynamics of the data.
In one of the most popular works in this direction, \citet{Schmidt2009} used symbolic regression to discover Hamiltonians, Lagrangians, and conservation laws for various physical systems, thus demonstrating its potential in scientific discovery.

Symbolic regression is often implemented through genetic programming, which searches through the space of mathematical expressions using evolutionary algorithms \cite{Koza1994}. 
The equations are pieced together through basic building blocks known as \textit{primitives}, which may include constants and simple functions (e.g. addition, multiplication, sine). 
However, these approaches do not typically scale well to high-dimensional problems and often require numerous hand-built heuristics and rules to ensure that the equation is simple to interpret and viable as a model.

There have been numerous approaches at introducing the power of deep learning into symbolic regression to enable learning equations in more complex settings. For example, AI-Feynman checks for a number of physics-inspired invariances and symmetries using both hand-built rules and neural networks to simplify the data \cite{udrescu2020ai}. SINDy (Sparse Identification of Nonlinear Dynamical systems) \cite{brunton2016discovering}, a system for discovering the governing equations of dynamical systems, has been combined with neural network to enable discovery on high-dimensional dynamical systems \cite{champion2019data}. PDE-Net 2.0 incorporates a symbolic network to discover partial differential equations (PDEs) using convolutional networks with constrained filters \cite{long2019pde}. \citet{lu2021discovering} incorporate a symbolic network with an encoder network to discover differential equation systems from partial observations. \citet{cranmer2020discovering} performs traditional symbolic regression on graph neural network weights after training in a 2-step process to discover the dynamics of many-body systems.

In particular, a neural network architecture called the equation learner (EQL) was proposed to perform symbolic regression, which takes a fully-connected neural network and replaces the activation functions with the primitive functions \cite{Martius2016,sahoo2018learning}. 
\citet{kim2020integration} expands upon the EQL network by integrating it with other deep learning architectures (including convolutional networks and recurrent networks) such that the entire model can be trained end-to-end through backpropagation, thus enabling symbolic regression on complex datasets including high-dimensional and dynamical systems where the relevant parameters may not be known ahead of time. 
Costa et al. \cite{costa2020fast} extends the EQL for recursive programs, implicit functions, and image classification.

One type of complexity we explore in this work are datasets described by parametric equations in which the underlying equation structure may stay the same but coefficients may vary along one or more dimensions.
PDEs are ubiquitous in describing the dynamics of many systems, but even the most simple settings can require varying coefficients.
For example, solving for electromagnetic modes or electron wavefunction in a material requires solving Maxwell's equation with spatially-varying permittivity or the Schr\"{o}dinger equation with varying potential \cite{griffiths_introduction_2018}, respectively.
Parameters may be influenced or even controlled by external factors that are not captured in the data \cite{zhang2002recurrent}.
The nonlinear Schr\"{o}dinger equation with varying coefficients has found applications in describing Bose-Einstein condensates \cite{yan2009exact}.
Additionally, the varying coefficients may change in complex ways that are not easily expressible symbolically, which would result in standard symbolic regression tools failing to discover interpretable equations.
Various approaches have been proposed to discover specifically parametric \textit{PDEs}, including group sparsity combined with SINDy \cite{rudy2019}, genetic algorithms combined with averaging over local windows \cite{xu2021deep}, and linear regression with kernel smoothing over adjacent coefficients \cite{luo2021ko}.

In this work, we propose to discover a more general class of parametric \textit{equations} where the equation structure is constant but the coefficients may vary in complex ways. We extend the approach from \citet{kim2020integration} and enable neural network-based symbolic regression on datasets governed by parametric equations. To this end, we propose two novel architectures: the stacked EQL network (SEQL) and the hyper EQL network (HEQL). We demonstrate our method on various analytic equations, PDEs, and a dataset consisting of 1D images of particles. In the last example, we combine the architectures with a convolutional neural network to analyze the images and demonstrate symbolic regression on high-dimensional datasets.

\section{Background}

\subsection{Equation Learner (EQL) Network}

The EQL network is a neural network architecture that can perform symbolic regression by replacing the nonlinear activation functions with primitive functions. The architecture was initially proposed in \cite{Martius2016,sahoo2018learning} and further expanded in \cite{kim2020integration}. In Section \ref{sec:base} we briefly review the base EQL architecture for symbolic regression, while more details can be found in ref. \cite{kim2020integration}. We also propose several modifications to the EQL network that improve its training behavior. In Sections \ref{sec:stack-architecture} and \ref{sec:para-architecture} we propose 2 variants of the EQL architecture that can discover parametric equations. 
Note that our discussion and notation below assume that the coefficients are parameterized with respect to \textit{time} as this provides a convenient intuition applicable to many systems. However, the parameterization could also be with respect to other quantities (e.g. space).

\subsubsection{Base Architecture} \label{sec:base}

\begin{figure*}
    \centering
    \includegraphics[width=0.75\textwidth]{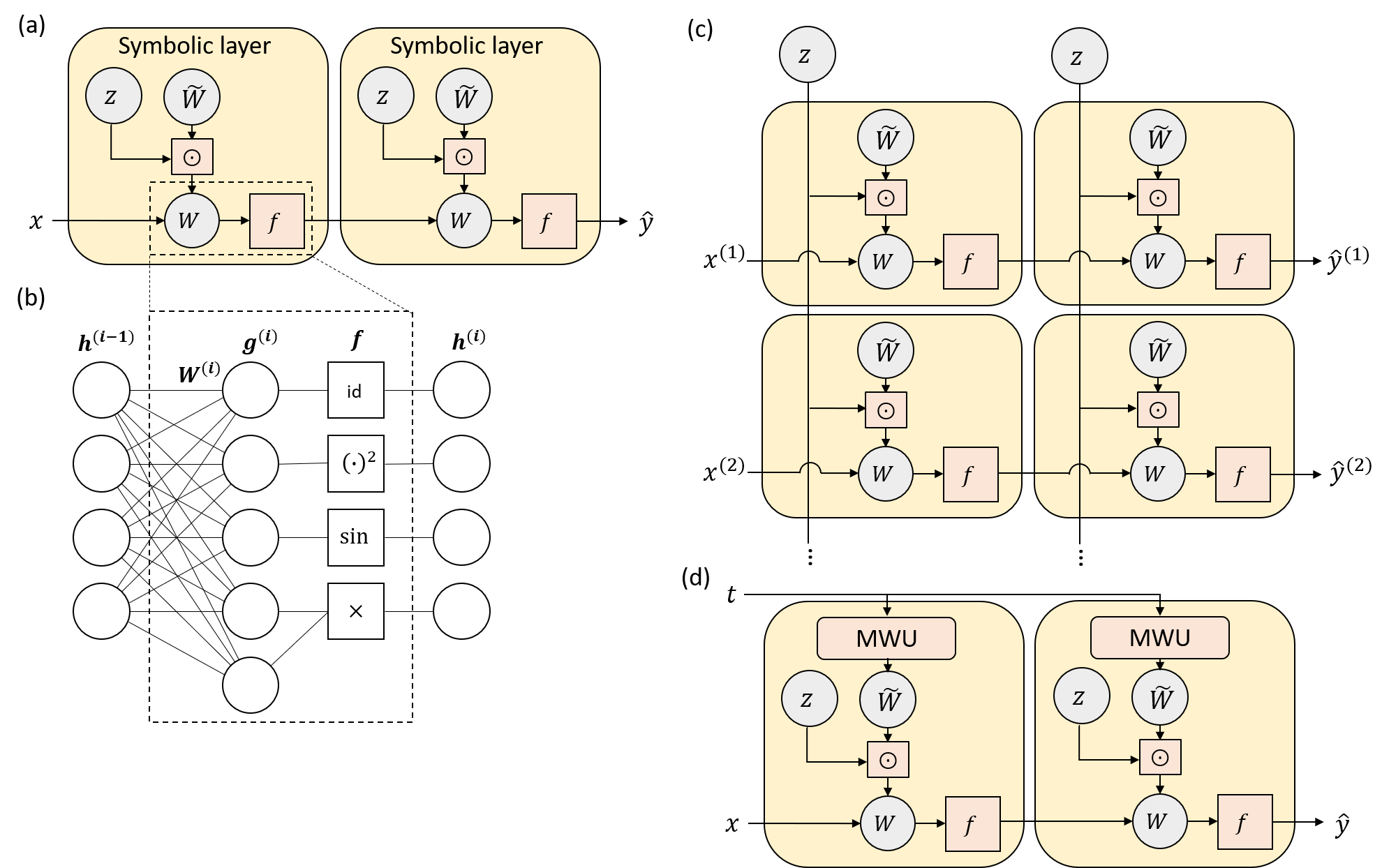}
    \caption{\textbf{Equation Learner (EQL) Architectures and Variants for Parametric Equations.}
    (a) Architecture of the base EQL network with relaxed $L_0$ regularization. 
    The weights $\mathbf{W}$ are re-parameterized as an element-wise product of the gate variables $\mathbf{z}$ and the weight values $\tilde{\mathbf{W}}$.
    (b) The core of the symbolic layer, where the activation functions consist of the primitive functions for symbolic regression, where each element may contain a different primitive function and primitive functions may take multiple inputs.
    (c) Architecture of the stacked EQL (SEQL) network. Note that the indexing $x^{(j)}$ is for the time step. Each horizontal row represents an EQL network for each time step. The gate $\mathbf{z}$ is shared across time steps.
    (d) Architecture of the hyper EQL (HEQL) network. 
    Note that in all schematics, the final (linear) layer is omitted for visual simplicity.
    }
    \label{fig:architecture}
\end{figure*}

The EQL network architecture closely resembles a fully-connected neural network in which the output $\mathbf{h}^{(i)}$ of the $i^{\mathrm{th}}$ layer can be described as 
\begin{align}
\mathbf{g}^{(i)} & = \mathbf{W}^{(i)}\mathbf{h}^{(i-1)} \label{eq:base1} \\
\mathbf{h}^{(i)} & = f\left(\mathbf{g}^{(i)} \right) \label{eq:base2}
\end{align}
\noindent where $\mathbf{W}$ is a weight matrix of the $i^{\mathrm{th}}$ layer, $f$ is the non-linear activation function, $\mathbf{g}^{(i)}$ is a vector of the pre-activation units, and $\mathbf{h}_0=\mathbf{x}$ is the input data. A schematic of a single layer is shown in Figure \ref{fig:architecture}(b). In regression tasks, the activation function for the final layer is typically omitted, so the output of the neural network with $L$ hidden layers is $y=\mathbf{W}^{(L+1)}\mathbf{h}^{(L)}$. 

While conventional neural networks typically use activation functions such as ReLU or sigmoid, the EQL network uses a set of functions that correspond to primitive functions in symbolic regression, which represent the building blocks for more complex equations. As shown in Figure \ref{fig:architecture}(b), each component of $
\mathbf{g}$ may go through a different primitive function, and a primitive function may also take multiple inputs (e.g., multiplication). The primitive functions may be duplicated to reduce the sensitivity of training to the random initialization. The network is trained using the same techniques as conventional neural networks, i.e. stochastic gradient descent, and once it is trained, the discovered equation can simply be read off of the weights.

\subsubsection{Sparsity}

To ensure the interpretability of symbolic regression, we need to force the system to learn the simplest expression that describes the data. In genetic programming-based approaches, this is typically done by limiting the number of terms in the expression. For the EQL network, we enforce this through the use of sparsity regularization on the network weights such that as many of the weights are set to 0 as possible. While Kim et al. \cite{kim2020integration} primarily use a smoothed $L_{0.5}$ regularization, in this work we use a relaxed form of $L_0$ regularization \cite{Louizos2017}. We briefly review the details here, and refer the reader to refs. \cite{kim2020integration} and \cite{Louizos2017} for more details.

The weights of the neural network are reparameterized as
\[\mathbf{W}=\mathbf{\tilde{W}}\odot \mathbf{z}\]
\noindent where $\mathbf{z}$ has the same dimensions as $\mathbf{W}$ and can be interpreted as a gate variable, and the multiplication is component-wise. Ideally each element of $\mathbf{z}$ is a binary ``gate" such that $z\in\{0,1\}$. However, this is not differentiable and so we allow $z$ to be a stochastic variable drawn from the hard concrete distribution \cite{Louizos2017}:
\begin{gather*}
u\sim\mathcal{U}(0,1) \\
s=\text{sigmoid}\left(\left[\log u - \log (1-u) + \log \alpha\right] / \beta \right) \\
\bar{s}=s(\zeta-\gamma)+\gamma) \\
z = \min(1, \max(0, \bar{s}))
\end{gather*}
\noindent where $u$ is a random variable drawn from the uniform distribution $\mathcal{U}$, $\alpha$ is a trainable variable that describes the location of the hard concrete distribution, and $\beta, \zeta, \gamma$ are hyperparameters that describe the distribution. The random variable $s$ is distributed as a binary concrete distribution, which is a continuous relaxation of a binary random variable \cite{maddison2016concrete}. Finally, the distribution is stretched out to the $(\gamma, \zeta)$ interval and ``folded'' to delta peaks at 0 and 1 to achieve the hard-concrete distribution.

In the case of binary gates, the regularization penalty would simply be the sum of $\mathbf{z}$ (i.e., the number of non-zero elements in $\mathbf{W}$). However, in the case of the hard concrete distribution, an analytical form for the expectation of the regularization penalty over the distribution parameters can be calculated \cite{Louizos2017}. The sparsity regularization loss is then
\[\mathcal{L}_R=\sum_{j} \text{sigmoid}\left(\log\alpha_{j} - \beta \log \frac{-\gamma}{\zeta} \right) \]
where $j$ is indexing through all of the weight components. While Louizos et al. \cite{Louizos2017} applies group sparsity to the rows of the weight matrices with the goal of computational efficiency, we apply parameter sparsity (to individual elements) with the goal of simplifying the expression in symbolic regression.

The advantage of $L_0$ regularization is that it enforces sparsity without placing a penalty on the magnitude of the weights by placing a penalty on the expected number of non-zero weights. Additionally, it lends itself to a straightforward definition of group sparsity across time-steps as we will see in Section \ref{sec:stack-architecture}. In our experiments, we use the hyperparameters for the $L_0$ regularization suggested by ref. \cite{Louizos2017}. 

\subsubsection{Skip Connections}

In this work, we also add skip connections to the EQL network to introduce an inductive bias towards simpler equations while simultaneously enabling the discovery of more complex equations. The most well-known type of skip connections were introduced in ResNets, which take the output of a layer and add it to the layer ahead with the goal of allowing gradient information to efficiently propagate though many layers and enabling extremely deep architectures \cite{he2016deep}. While these would be feasible to implement in the EQL network, they would increase the complexity of the equation as information flows through the network. In contrast, we turn to the skip connections introduced by DenseNets which concatenates, rather than sums, the output of the previous layer with that of the next layer \cite{huang2017densely}. More specifically, we modify Equation \ref{eq:base2} as:
\begin{equation}
    \mathbf{h}^{(i)} =  \left[ f\left(\mathbf{g}^{(i)} \right) ; \mathbf{h}^{(i-1)} \right]
\end{equation}

Skip connections introduce a slight inductive bias towards learning simpler functions, since functions can route ``directly'' to the output without needing to go through the identity primitive function of successive layers. Additionally, skip connections minimize instabilities during training that can arise as a result of gradients exploding as they pass through the primitive functions. Thus, skip connections allow us to train EQL networks with more layers, which in turn can enable learning more complex equations.

\subsection{Parametric Equations} \label{sec:parametric-intro}

\begin{figure*}
    \centering
    \includegraphics[width=0.8\textwidth]{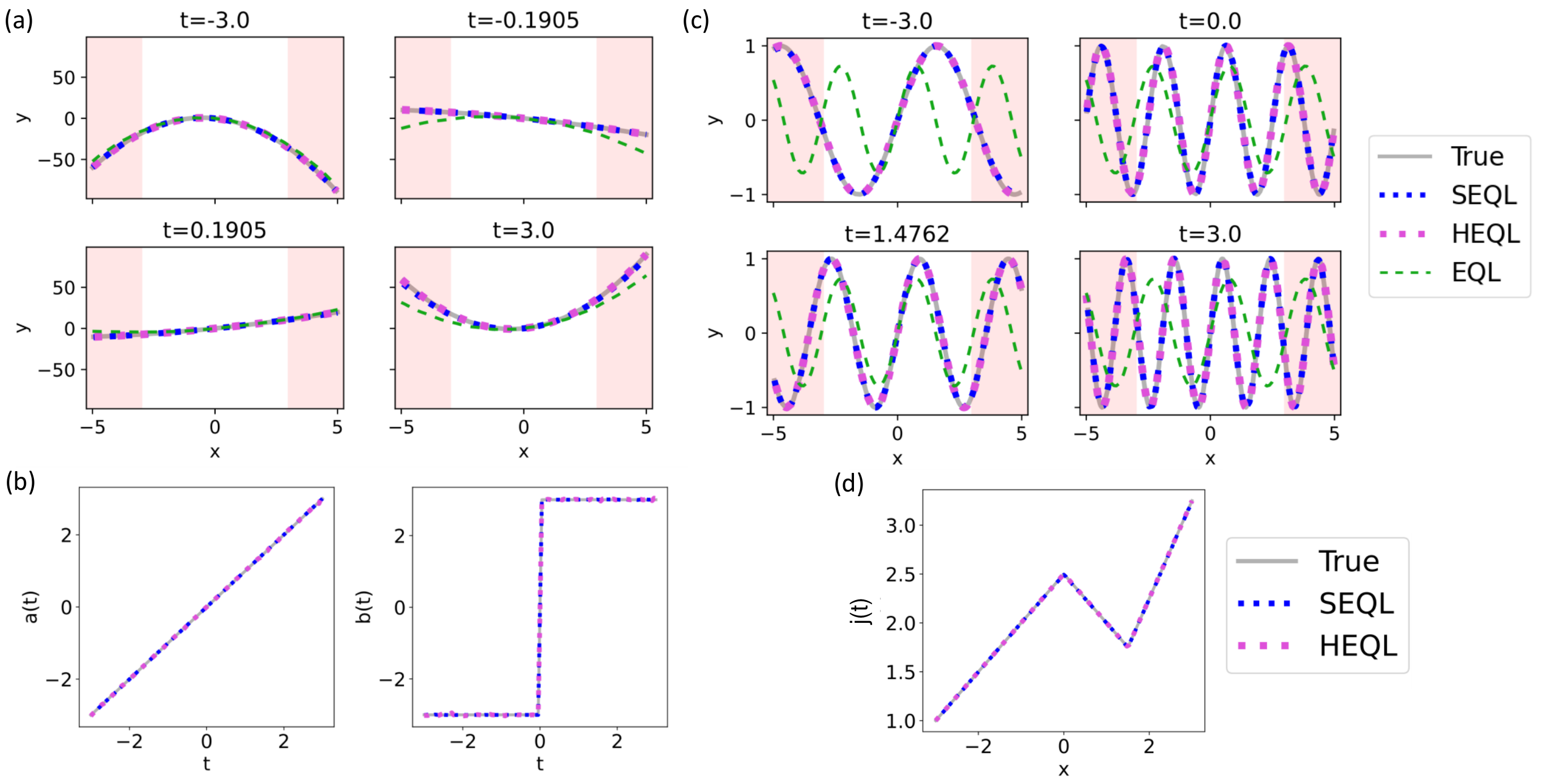}
    \caption{\textbf{Learning parametric equations.}
        (a, b) Learning the function $f_1$ which contains a discontinuity at $t=0$.
        (c, d) Learning the function $f_2$ which corresponds to a sinusoid with a frequency that varies non-smoothly as a function of $t$.
        (a, c) Predictions after training the EQL, SEQL, and HEQL networks in the range $-3 < x < 3$ for various values of $t$. Values outside of this range (highlighted in red) are extrapolated.
        (b, d) Learned functions for the varying coefficents.
        }
    \label{fig:analytic}
\end{figure*}

In this work we focus specifically on learning equations where the \textit{structure} of the equation (i.e., terms and operators) remains constant but the value of the numerical coefficients or constants may vary over the dataset. 
Prior works have focused specifically on learning partial differential equations (PDEs) where the differential terms are fixed but the coefficients vary over space or time, termed \textit{parametric} PDEs \cite{rudy2019,xu2021deep,luo2021ko}. 
Here we generalize symbolic regression to parametric \textit{equations}.
The ability to learn parametric equations may greatly expand the scope of symbolic regression, especially in cases where the coefficients may vary over the dataset in arbitrarily complex ways that are difficult to express symbolically. 
Such behavior would greatly impede the performance of traditional symbolic regression approaches that attempt to find the simplest equation describing the dataset. 
To illustrate this, we train the EQL network on two simple parametric equations listed in Table \ref{tab:analytic-eql}.

\begin{table}
    \centering
    \caption{Learned equations by the EQL network after training on simple parametric equations.}
    \begin{tabular}{ll}
        \myalign{c}{True} & \myalign{c}{EQL}  \\ \hline
        $f_1=tx^2+3\sgn(t)x$ & \begin{tabular}{@{}l@{}}
                    $0.02 x^2 - 0.14 t + 1.86 x t + 1.01 x^2 t$  \\
                    $\hphantom{~~} + 0.03 t^2 - 0.01 x^2 t^2 - 0.02 t^3 - 0.01 x t^3 $\\
                    $\hphantom{~~} +(0.09 - 1.80 x + 0.01 x^2) \sin(1.91t)$ \\
                    $\hphantom{~~} + 0.77 x \sin(3.71 t) + 0.32 \sin(3.22 x t)$ \\
                    $\hphantom{~~} - 0.10 $
                    \end{tabular}
             \\
        $f_2=\sin(j(t)x)$ & $0.720\sin(2.05x)$ \\
    \end{tabular}
    \label{tab:analytic-eql}
\end{table}

The first function is a parabolic curve described as:
\begin{equation*}
f_1(x, t) = tx^2 + 3\sgn(t)x \\
\end{equation*}
where $\sgn$ is the \textit{sign} function (also known as the \textit{signum} function):
\begin{equation*}
\sgn(t) = 
    \begin{cases}
        -1 & \text{if } t < 0 \\
        0 & \text{if } t = 0 \\
        1 & \text{if } t > 0
    \end{cases}.
\end{equation*}
The function notably contains a discontinuity at $t=0$ and thus cannot be described in terms of smooth functions. 

As seen in Table \ref{tab:analytic-eql}, the EQL network learns an overly complicated and incorrect equation with over a dozen terms that likely signify its attempt to fit the discontinuity. It is difficult to interpret and thus fails the goal of discovery. Additionally, the results of fitting the EQL network to the data are shown in Figure \ref{fig:analytic}(a). While the EQL network seems to fit reasonably well inside the training regime (in the range $-3 < x < 3$), it fails to extrapolate well since it has not learned the correct equation. (In principle, the function $f_1$ is simple enough such that an EQL network with sigmoidal activation functions could approximate it with reasonable accuracy, but we choose this example to illustrate some of the difficulties of symbolic regression.)

In the second example, we look at a sinusoidal curve where the frequency varies non-smoothly as a function of time:
\begin{equation*}
    f_2(x, t) = \sin(j(t)x) 
\end{equation*}
where we have defined a ``jagged'' function:
\begin{equation*}
    j(t) =
    \begin{cases}
        0.5t + 2.5   & \text{if } t < 0 \\
        -0.5t + 2.5  & \text{if } 0 \leq t < 1.5 \\
        t + 0.25             & \text{otherwise}
    \end{cases}.
\end{equation*}
The jagged function $j(t)$ is illustrated in Figure \ref{fig:analytic}(d). As seen in Table \ref{tab:analytic-eql}, the EQL network is unable to learn the parametric form of the sinusoidal frequency, and thus fits poorly.

In contrast, the proposed architectures in Section \ref{sec:parametric} are able to learn the correct equations as shown in Table \ref{tab:analytic-results}, as we will discuss next.

\section{Parametric EQL Variants} \label{sec:parametric}

In this work, we propose two variants of the EQL network to learn parametric equations: the Stacked EQL (SEQL) and the Hyper EQL (HEQL).

\subsection{Stacked Architecture (SEQL)} \label{sec:stack-architecture}

The first extension we propose to analyze parametric equations is to train a separate EQL network for each time step, an architecture that we call the stacked EQL (SEQL) network. Suppose we have a dataset that is indexed by the time step $j$:
\begin{equation} \label{eq:stacked-dataset}
    \mathcal{D}=\left\{ \left\{ x^{(i, j)}, y^{(i, j)} \right\}_{i=1}^{N^{(j)}} \right\}_{j=1}^{N_t}
\end{equation} 
where $N_t$ is the number of time steps and $N^{(j)}$ is the number of data points in the $j$th time step (note that $N^{(j)}$ does not need to be constant across time steps). For layer $i$ of the SEQL network, we can construct $N_t$ separate weight matrices, $\left\{\mathbf{\tilde{W}}^{(i,j)}\right\}_{j=1}^{N_t}$, such that Equations \ref{eq:base1} and \ref{eq:base2} are modified as:
\begin{align}
\mathbf{g}^{(i,j)} & = \mathbf{W}^{(i,j)}\mathbf{h}^{(i-1,j)} \\
\mathbf{h}^{(i,j)} & = f\left(\mathbf{g}^{(i,j)} \right). \label{eq:seql}
\end{align}
In other words, we parameterize a separate EQL network for each time step, as shown in Figure \ref{fig:architecture}(c).

If we na\"ively train $N_t$ separate EQL networks, then it is possible that each network may learn a different equation in each time step. Additionally, each network would only see approximately $\frac{1}{N_t}$ of the total data, thus reducing data efficiency. To counteract this, we enforce that the different networks learn the same equation by implementing group sparsity through weight sharing of the gate variable $\mathbf{z}$. For the $i^{\mathrm{th}}$ layer of the $j^{\mathrm{th}}$ time step, we further modify Eq. \ref{eq:seql} as:
\begin{equation}
    \mathbf{h}^{(i,j)} = f\left( \left( \mathbf{\tilde{W}}^{(i,j)} \odot \mathbf{z}^{(i)}\right)\mathbf{h}^{(i-1,j)} \right)
\end{equation}
Note that $\mathbf{z}$ is not parameterized with respect to the time step $j$. For an architecture with $L$ hidden layers, there are $(L+1)\cdot N_t$ weight matrices and $L+1$ gate matrices.

Another modification we make to the architecture is weight regularization across time steps to introduce an inductive bias towards smoothness in the coefficients. We use $L_2$ regularization loss between adjacent time steps. Looking at just a single element $w_{k,l}$ of $\mathbf{W}$ in a single layer for notational simplicity, the inter-layer $L_2$ loss is simply
\begin{equation}
    L_{S,k,l}=\sum_{j=1}^{N_t-1}\left(w_{k,l}^{(j+1)}-w_{k,l}^{(j)}\right)^2
\end{equation}
and the total inter-layer regularization loss is
\begin{equation}
    \mathcal{L}_{S}=\sum_{i,k,l} L_{S,k,l}^{(i)}
\end{equation}
where $i$ indexes the layer. This regularization pushes coefficients in adjacent time-steps closer together and can more effectively counteract noisy datasets.

\subsection{Hyper EQL (HEQL) Architecture} \label{sec:para-architecture}

We also propose a second variant of the EQL network, the Hyper EQL (HEQL) network, in which the weights $\tilde{\mathbf{W}}$ are re-parameterized as a function of the varying coefficient, i.e., $\tilde{\mathbf{W}}(t)$.
While a number of models can be used to parameterize the weights, we use a fully-connected neural network as it is a flexible model that can fit arbitrary functions and can be trained with backpropagation, allowing the entire system to be trained end-to-end. We call this fully-connected neural network the meta-weight unit (MWU). The architecture is shown in Figure \ref{fig:architecture}(d).

This idea is similar to that of hypernetworks, in which a neural network is used to generate the weights of another neural network \cite{ha2016hypernetworks}. The general idea of using a network to parameterize or interact with the weights of another network has been most notably leveraged for meta-learning \cite{andrychowicz2016learning,munkhdalai2017meta,ravi2017optimization,hospedales2020meta}, and has also been applied to a variety of other architectures, including the Neural ODE \cite{chen2018neural} and HyperPINN \cite{de2021hyperpinn}. 

The HEQL has a separate MWU in each layer (including the linear output layer) which takes the parametric variable $t$ as an input and outputs the weight matrix $\mathbf{\tilde{W}}^{(i)}(t)$ for that layer. 
The gate variables $\mathbf{z}$ are not modified and are thus not a function of $t$. 
As a result, all of the ``time steps'' share the same sparsity regularization, thus avoiding the need for any further modifications to implement group sparsity.

The advantage of this architecture compared to the SEQL is that the HEQL does not replicate the EQL network for each time step, thus saving on computational memory especially for large $N_t$. The architecture can also make predictions on a continuous domain of $t$ and does not need require the data to align along a fixed grid in time, unlike the prior work on discovering parametric PDEs \cite{rudy2019,xu2021deep,luo2021ko}. More specifically, rather than viewing the dataset as Equation \ref{eq:stacked-dataset}, we have greater flexibility and can view the dataset as
\begin{equation}
    \mathcal{D}=\left\{ x^{(i)}, y^{(i)}, t^{(i)} \right\}_{i=1}^{N}.
\end{equation}

Although we do not explicitly regularize the functional space of the parametric coefficients, neural networks tend to generalize well despite typically being overparameterized, which is a topic of significant interest \cite{nakkiran2021deep,liu2020understanding,liu2022convnet,jakubovitz2019generalization}. In practice, this means that the predictions of neural networks for regression tasks tend to be smooth, and so the function of the parametric coefficient will also tend to be smooth.

\section{Results} \label{sec:results}

We now look at several different problem settings with parametric quantities that can be analyzed by our system. For simplicity, we highlight some of the results here, and the remainder can be found in the appendix. Section \ref{sec:analytic} demonstrates some simple benchmarks to highlight the aspects of learning parametric equations. Section \ref{sec:pde} shows results on PDE datasets taken from other works. Finally, section \ref{sec:spring} presents results on 1D images of a spring system to demonstrate the ability to perform symbolic regression on higher-dimensional systems.

Our datasets and code are made publicly available at \url{https://github.com/samuelkim314/parametric-eql}.

\subsection{Analytic Expressions} \label{sec:analytic}

\begin{table*}
    \centering
    \caption{Results for training on parametric analytic expressions. Learned equations are extracted for various values of $t$.}
    \begin{tabular}{crrrr}
        & & & \multicolumn{2}{c}{Learned equations}\\
        \cmidrule(lr){4-5}
        Benchmark  & \myalign{c}{$t$} & \myalign{c}{True} & \myalign{c}{SEQL} & \myalign{c}{HEQL} \\ \hline
        \multirow{4}{*}{$f_1(t,x)$} 
                & $-2.619$ & $-2.619x^2-3x$ & $-2.621x^2-2.995x-0.045$ & $-2.628x^2-3.017x-0.026$ \\
            & $-1.095$ & $-1.095x^2-3x$ & $-1.096x^2-3.003x-0.024$ & $-1.106x^2-3.005x-0.011$ \\
            & $0.381$ & $0.381x^2+3x$ & $0.383x^2+3.001x+0.006$ & $0.386x^2+3.013x + 0.005$ \\
            & $1.905$ & $1.905x^2+3x$ & $1.908x^2+3.000x+0.025$ & $1.909x^2+3.020x+0.031$\\ \hline
        \multirow{4}{*}{$f_2(t,x)$}
            & $-2.619$ & $\sin(1.191x)$ & $0.999\sin(1.191x)$ & $1.000\sin(1.190x)$ \\
            & $-1.095$ & $\sin(1.953x)$ & $0.999\sin(1.952x)$ & $1.000\sin(1.952x)$ \\
            & $0.381$ & $\sin(2.310x)$ & $0.997\sin(2.310x)$ & $1.000\sin(2.309x)$ \\
            & $1.905$ & $\sin(2.155x)$ & $0.999\sin(2.155x)$ & $1.000\sin(2.155x)$\\
    \end{tabular}
    \label{tab:analytic-results}
\end{table*}

To verify the ability of the SEQL and HEQL networks to discover parametric equations, we benchmark the networks on the analytical expressions discussed in Section \ref{sec:parametric-intro} and listed in Table \ref{tab:analytic-eql}. While we train the networks on data drawn from the domain $x\in[-3,3]$, we test the networks on a wider domain $x\in[-5,5]$ to evaluate extrapolation performance. In Appendix \ref{app:results-analytic}, we also provide benchmarks for additional analytic expressions.

Figure \ref{fig:analytic} shows the results for learning $f_1$ and $f_2$ using the SEQL and HEQL networks. 
The true function and the predicted function are plotted for various values of $t$. In all cases, the SEQL/HEQL predictions are visually indiscernible from the true function in both the training regime and the test regime, demonstrating that the architectures are able to extrapolate. 
As in the case of the original EQL network, the learned equations can be extracted from the trained network by simply processing the learned weights with software for symbolic mathematics.
In particular, we use SymPy, a Python package for symbolic mathematics, to simplify the resulting expression \cite{sympy}.
For clarity, we also omit negligible terms (i.e., those with coefficient magnitudes $<0.01$) in the final expression.
The extracted equations found by the SEQL and HEQL for the parametric function $f_1$ and $f_2$ at various time steps are shown in Table \ref{tab:analytic-results}.
The mean squared error (MSE) on the test datasets can be found in Table \ref{tab:analytic-benchmark-mse} in Appendix \ref{app:results-analytic}.

Upon inspection of the extracted equations over multiple time steps for learning $f_1$, we see that the architectures has successfully discovered the function $\hat{f}_1=a(t)x^2+b(t)x+\epsilon(t)$ where $a(t)$ and $b(t)$ are the varying coefficients and $\epsilon$ is a small number that can either be eliminated with further training or ignored upon inspection.
The predicted parametric coefficients $a(t)$ and $b(t)$ match the true coefficients extremely closely, as seen in Figure \ref{fig:analytic}(b). 
Note that the SEQL/HEQL networks are able to learn the discontinuous $\sgn$ function without any apparent smoothing at $t=0$. 
Discontinuous coefficients would be difficult to learn using other methods for parametric equations that rely on local averaging \cite{xu2021deep} or smoothing \cite{luo2021ko}.
This also contrasts with the original EQL network which was unable to learn the parametric equation as shown in Table \ref{tab:analytic-eql}.

For learning the sinusoidal function $f_2$, both the SEQL and HEQL networks have learned the equation $\hat{f}_2=\sin(a(t)x)$ as seen in Table \ref{tab:analytic-results} where $a(x)$ is plotted in Figure \ref{fig:analytic}(d). Again, the predictions match the true function extremely well across time steps and outside of the training regime. Although sinusoidal functions are typically difficult to learn through linear regression techniques, the SEQL and HEQL networks are able to learn this function across multiple spatial frequencies. 

Note that because the varying coefficient is inside the $\sgn$ and $\sin$ functions for $f_1$ and $f_2$, respectively, other methods for learning parametric equations such as those proposed in Refs. \cite{luo2021ko} or \cite{brunton2016discovering} that rely on linear regression techniques would not be able to discover these types of equations. 
In contrast, the multi-layer architecture of the SEQL and HEQL networks allow for the varying coefficient to be inside nested functions, enabling discovery of much more complex parametric equations.

Results for additional benchmark equations are listed in Appendix \ref{app:results-analytic}. 
Interestingly, there is no clear trend on whether the SEQL or HEQL tends to perform better.

\subsection{Partial Differential Equations (PDEs)} \label{sec:pde}

\begin{figure}
    \centering
    \includegraphics[width=0.959\columnwidth]{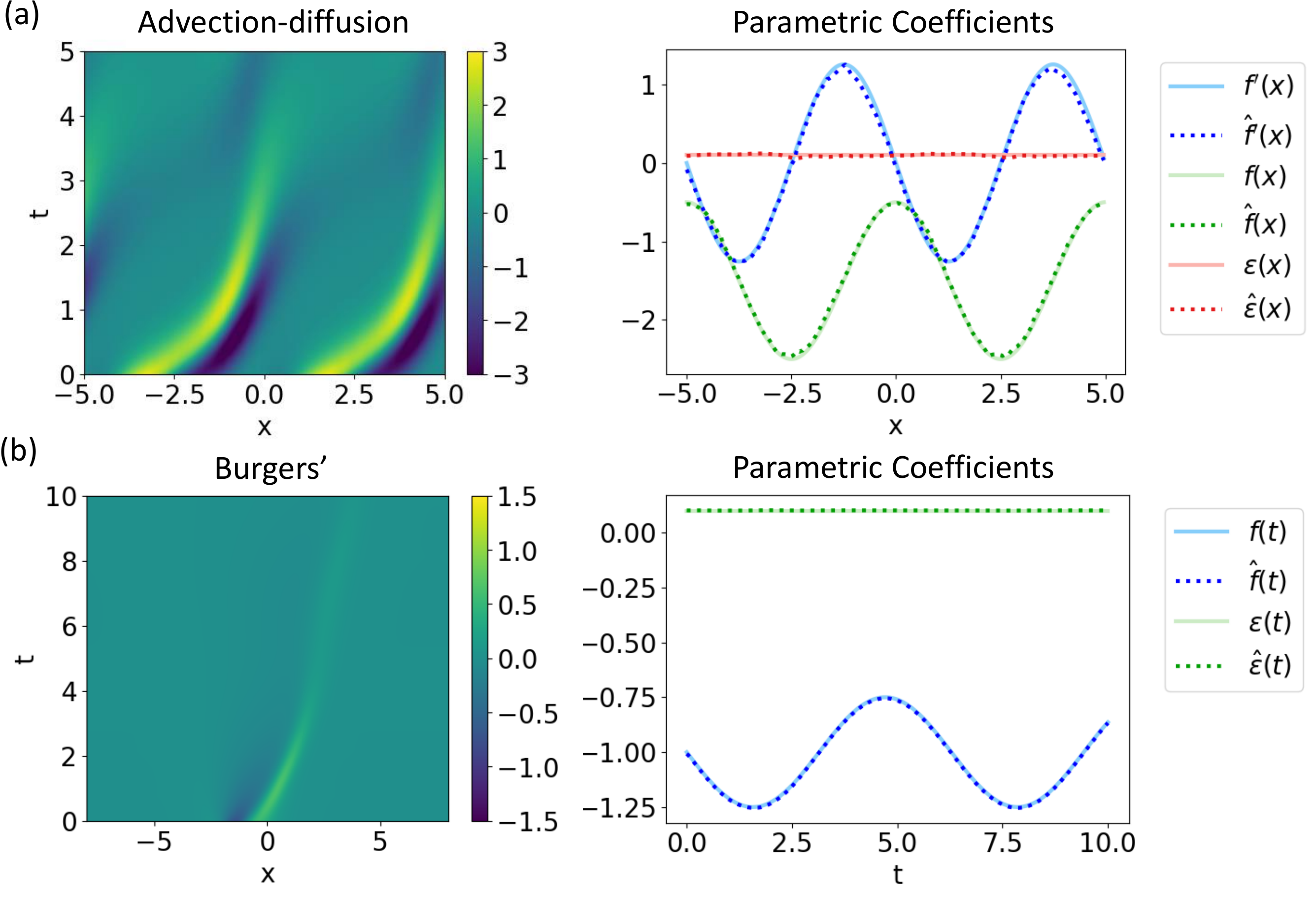}
    \caption{Results for learning (a) the advection-diffusion equation using the HEQL network, and (b) Burgers' equation using the SEQL network. Left-hand plots show the predicted values of $u_t$ and the right-hand plots illustrate the varying coefficient functions.}
    \label{fig:pde-results}
\end{figure}

\begin{table*}
\centering
\caption{MSE after training on PDE datasets.}
\begin{tabular}{lcccccc}
                                      & \multicolumn{3}{c}{$u_t$ MSE}                   & \multicolumn{3}{c}{Coefficient MSE}                      \\
                                      \cmidrule(lr){2-4} \cmidrule(lr){5-7}
Benchmark        & SINDy         & SEQL          & HEQL          & SINDy         & SEQL          & \multicolumn{1}{l}{HEQL} \\ \hline
Advection-diffusion, with cross terms & $\mathbf{2.34\times10^{-7}}$ & $2.87\times10^{-5}$ & $1.99\times10^{-5}$ & $3.99\times10^{-2}$ &  $2.99\times10^{-3}$ &    $\mathbf{1.44\times10^{-3}}$   \\
Advection-diffusion, no cross terms   & $\mathbf{2.34\times10^{-7}}$ & $2.12\times10^{-5}$ & $1.57\times10^{-5}$ & $1.73\times10^{-2}$ & $2.30\times10^{-3}$ & $\mathbf{1.37\times10^{-3}}$            \\
Burgers', with cross terms            & $\mathbf{5.11\times10^{-8}}$ & $5.66\times10^{-6}$ & $5.54\times10^{-6}$ & $\mathbf{3.95\times10^{-7}}$ &  $8.81\times10^{-6}$ &   $9.83\times10^{-6}$   \\
Burgers', no cross terms              & $2.30\times10^{-4}$ & $\mathbf{2.55\times10^{-7}}$ & $8.02\times10^{-6}$ & $1.73\times10^{-3}$ & $\mathbf{8.41\times10^{-6}}$ & $4.44\times10^{-5}$           
\end{tabular}
\label{tab:pde-mse}
\end{table*}

Next, we investigate learning partial differential equations (PDEs) with varying coefficients from data. In this setting, a quantity of interest $u(x, t)$ can be defined by a function of its partial derivatives (e.g. $u_t$, $u_x$, $u_xx$) and a parametric dependence on time, $\mu(t)$:
\[
    u_t = N(u, u_x, u_xx, ..., \mu(t))
\]
where $N$ is the evolution function that we wish to learn. For notational convenience, we drop the explicit dependence of $u$ on $x$ and $t$.

Prior works in discovering parametric equations have focused on the setting of PDEs \cite{rudy2019,xu2021deep,luo2021ko}, as PDEs are ubiquitous in describing dynamics in a variety of fields. For ease of comparison, we benchmark our architectures on two of the datasets provided by \citet{rudy2019}: the advection-diffusion equation and Burgers' equation with varying coefficients. The partial differential terms (e.g. $u_x, u_{xx}$) are pre-computed from the dataset and concatenated with the input $u$. 

We note that SINDy is not able to automatically calculate cross terms (e.g. $uu_x$) and so the cross terms were also pre-computed and fed into SINDy in the original work \cite{rudy2019}. We label this approach as ``with cross terms'' in Table \ref{tab:pde-mse}. In contrast, the SEQL and HEQL architectures are able to automatically discover cross terms as necessary, and so we also carry out experiments that omit the cross terms in the input, labelled ``no cross terms.''

\subsubsection{Advection-Diffusion Equation}

The advection-diffusion equation describes numerous physical transport systems and has been applied to describe the movement of pollutants, reservoir flow, heat, and semiconductors. We use an adaptation of the equation that includes a spatially-dependent velocity field, as in \cite{rudy2019}:
\begin{gather}
    u_t = f'(x)u + f(x)u_x + \epsilon u_{xx}\text{.}
\end{gather}
where $f(x)=-1.5+\cos\left(\frac{2\pi x}{5}\right)$ and $\epsilon=0.1$. Note that the parametric quantities vary with respect to space rather than time. Thus, for our experiments, we modify the SEQL and HEQL architectures to parameterize the varying coefficient with respect to space (in practice this simply involves relabeling the dataset).

\begin{table*}
    \centering
    \caption{Learned equations on select $x$ or $t$ values (depending on the varying parameter) for the PDE datasets without cross terms.}
    \begin{tabular}{lrrrr}
        \myalign{c}{Benchmark} & \myalign{c}{$x$ or $t$} & \myalign{c}{True} & \myalign{c}{SEQL} & \myalign{c}{HEQL} \\ \hline
        \multirow{4}{*}{Advection-diffusion equation}
        & $-4.375$ & $-0.89u-0.79u_x+0.10u_{xx}$ & $-0.86u-0.77u_x+0.11u_{xx}$ &  $-0.86u-0.76u_x+0.11u_{xx}$ \\
        & $-1.875$ & $0.89u-2.21u_x+0.10u_{xx}$ & $0.83u-2.14u_x+0.07u_{xx}$ & $0.86u-2.16u_x+0.09u_{xx}$\\
        & $0.625$ & $-0.89u-0.79u_x+0.10u_{xx}$ & $-0.86u-0.77u_x+0.11u_{xx}$ & $-0.86u-0.77u_x+0.11u_{xx}$ \\
        & $3.125$ & $0.89u-2.21u_x+0.10u_{xx}$ & $0.83u-2.14u_x+0.07u_{xx}$ & $0.88u-2.17u_x+0.10u_{xx}$ \\ \hline
        \multirow{4}{*}{Burgers' equation}
        & $0.627$ & $-1.15uu_x+0.10u_{xx}$ & $-1.15uu_x+0.10u_{xx}$ & $-1.16uu_x+0.10u_{xx}$ \\
        & $3.137$ & $-1.00uu_x+0.10u_{xx}$ & $-1.00uu_x+0.10u_{xx}$ & $-1.01uu_x+0.10u_{xx}$ \\
        & $5.647$ & $-0.85uu_x+0.10u_{xx}$ & $-0.86uu_x+0.10u_{xx}$ & $-0.85uu_x+0.10u_{xx}$ \\
        & $8.157$ & $-1.24uu_x+0.10u_{xx}$ & $-1.24uu_x+0.10u_{xx}$ & $-1.25uu_x+0.10u_{xx}$ \\
    \end{tabular}
    \label{tab:pde-eqs}
\end{table*}

Table \ref{tab:pde-mse} lists the MSE for both the predicted $u_t$ as well as the learned coefficient functions.
Unsurprisingly, SINDy achieves the lowest MSE on $u_t$ since it reduces the problem to a linear system that can be solved efficiently.
Interestingly, SEQL and HEQL achieve better predictions on the coefficient functions, which is perhaps due to the implicit regularization in the architectures.
Without cross terms, the SEQL and HEQL achieve an even lower error, which may be due to the reduced dimensionality of the input.
Table \ref{tab:pde-eqs} shows the equations that the SEQL and HEQL have learned after training for select values of $x$.
Both networks have learned an equation of the form $\hat{u}_t=\hat{f}'(x)u+\hat{f}(x)u_x+\hat{\epsilon}(x)u_{xx}$, and have thus successfully discovered the equation structure. 
The predicted $\hat{u}_t$ along with the learned parametric coefficients (i.e. $\hat{f}'(x), \hat{f}(x), \hat{\epsilon}(x)$) are shown in Figure \ref{fig:pde-results}(a) for the HEQL network, which match the actual values very closely. 
Results for the SEQL (not shown) are visually very similar.

\subsubsection{Burgers' Equation}

Burgers' equation is an important differential equation originally proposed to model turbulent flow that has also been applied to other processes such as traffic flow and boundary layer behavior. Here we analyze Burgers' equation with an oscillating coefficient for the non-linear term, as in \cite{rudy2019}:
\begin{gather}
    u_t = f(t)uu_x + \epsilon u_{xx}\text{.}
\end{gather}
where $f(t)=-\left(1+\frac{\sin(t)}{4}\right)$ and $\epsilon=0.1$. 

Note that this equation contains a cross term, $uu_x$. When including the cross term in the input, SINDy is correctly learn the equation and achieve a low error, as shown in Table \ref{tab:pde-mse}. 
However, when the cross term is omitted from the input, SINDy is unable to learn the correct equation and adds incorrect terms (i.e. $u_x$, $u_{xxx}$) to compensate. 
In contrast, the SEQL and HEQL are able to achieve low errors in both cases, and achieve the correct equation form as shown in Table \ref{tab:pde-eqs}. 
Figure \ref{fig:pde-results}(b) shows that the SEQL network is able to accurately predict the function and the parametric coefficients. 
Thus, our system is able to automatically learn these cross terms using the multiplication primitive, and more generally, can learn the form of a nonlinear PDE.

\subsection{Spring System} \label{sec:spring}

Finally, we demonstrate the ability of the parametric EQL networks to perform symbolic regression on structured, high-dimensional data by integrating our architectures with other deep learning architectures and training the entire model end-to-end.

We consider a dataset that consists of pairs of 1D images of point particles that interact through a spring-like force. The input data is a 1D grayscale  image with $64$ pixels which represents a 1D spatial domain $\psi\in[-4,4]$. Each image contains a single particle, represented by a Gaussian with mean centered at its position $\psi_i$ and a fixed variance of $0.1$. We look at two different targets for symbolic regression: the spring force
\begin{equation}F=-k(t)(\psi_2-\psi_1) \end{equation}
and the spring energy 
\begin{equation}E=\frac{k(t)}{2}(\psi_2-\psi_1)^2 \end{equation}
where $k(t)=\frac{5-t}{2}$. These are interpretable equations in that we know that the spring force and potential only depend on the spring constant, $k(t)$, and the distance between the two particles. The spring constant decreases over time, which we can imagine represents a spring degrading with use. The manner in which the spring degrades may or may not be analytical, and so we treat this as a parametric quantity.

\begin{figure}[tb]
    \centering
    \includegraphics[width=\columnwidth]{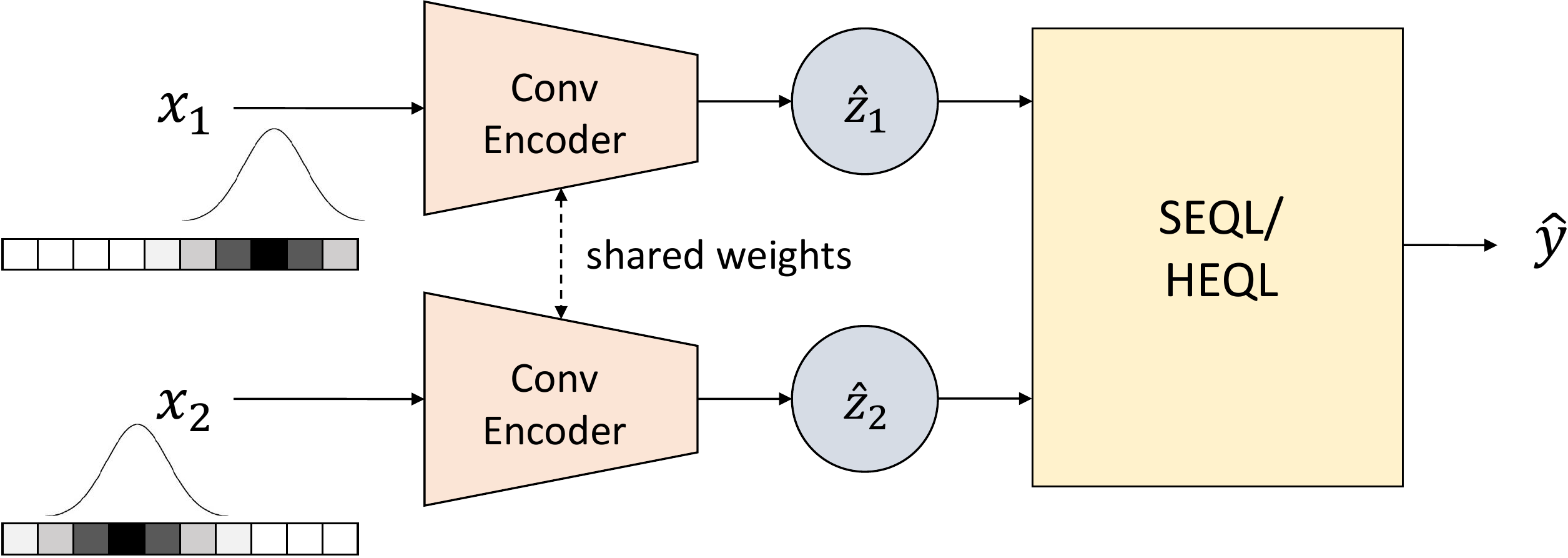}
    \caption{The combined architecture used for high-dimensional system tasks involving a convolutional encoder followed by an EQL network.}
    \label{fig:conv-architecture}
\end{figure}

To approach this problem, we use the architecture shown in Figure \ref{fig:conv-architecture}. Each image is fed into a separate encoder, where the two encoders share the same weights. The encoder consists of $2$ convolutional layers followed by $3$ fully-connected layers and a batch normalization layer. 
Each encoder outputs a single-dimensional latent variable (either $\hat{z}_1$ or $\hat{z}_2$) which are then fed into the parametric EQL network (which can be either the SEQL or the HEQL). 
The batch normalization layer serves to constrain the range of the latent variable so that the SEQL/HEQL network does not need to scale to arbitrarily-sized inputs when training end-to-end. The SEQL/HEQL network has a single scalar output, which is trained to match either the spring force or the spring energy. The entire network is trained end-to-end and is only shown the inputs and the output, but must learn an appropriate representation $\hat{z}_i$. While there are no constraints on the latent representation $\hat{z}_i$, we expect it to have a one-to-one mapping to the true position of the particle, $\psi_i$.

For all tests, $512$ training data points with $\psi_1,\psi_2\in[-3,3]$ were sampled for each of $128$ fixed values of $t\in[-3,3]$. To evaluate the extrapolation ability of these architectures, training data points were restricted to pairs with $|\psi_2-\psi_1|\leq4$, while no such restriction was imposed on testing data. In addition, we compare against a baseline test of a model consisting of the same encoder architecture with a dense ReLU network replacing EQL network. We call this baseline the \textbf{ReLU network}. 

\begin{figure}
    \centering
    \includegraphics[width=0.85\columnwidth]{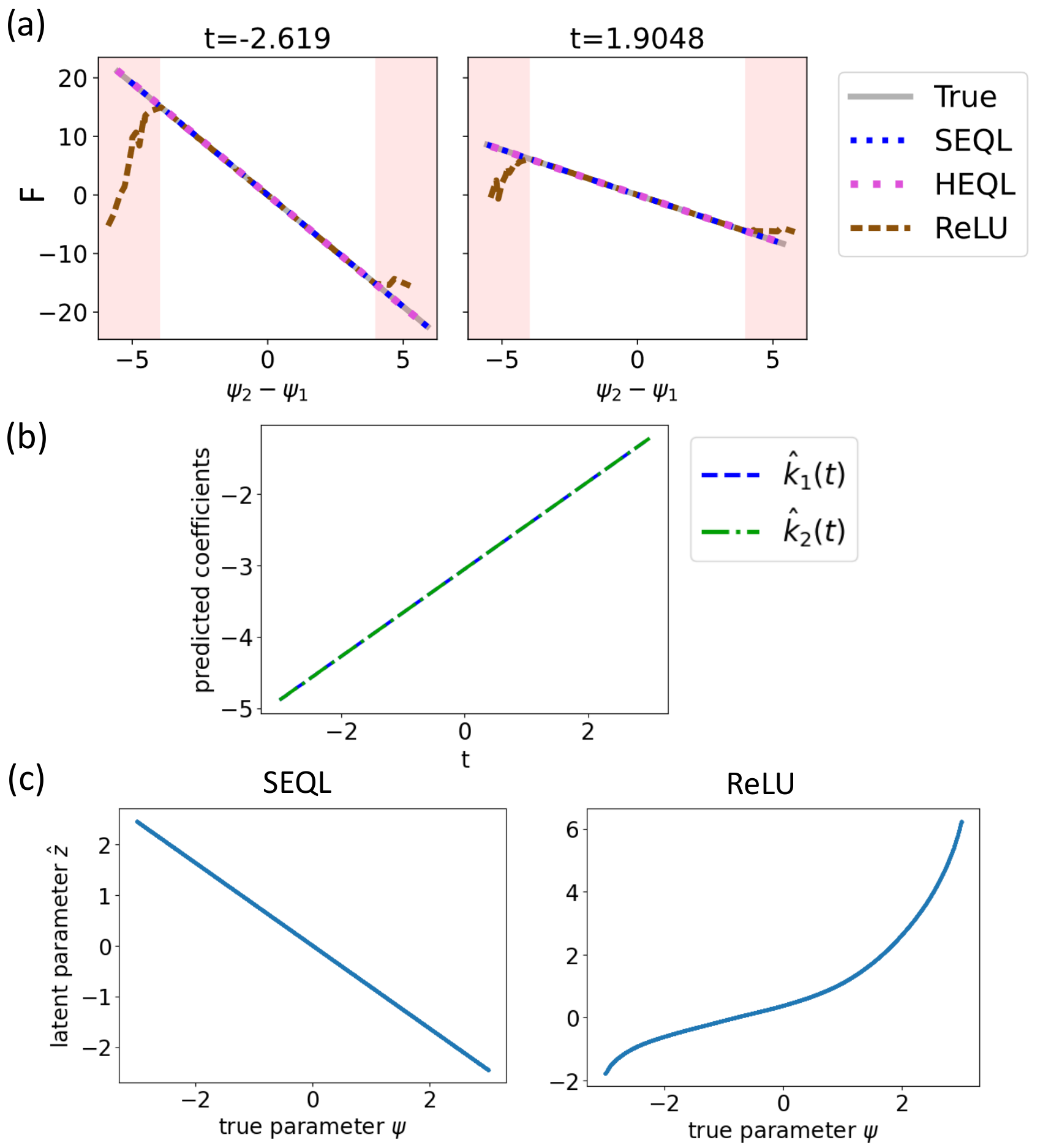}
    \caption{\textbf{Results for learning the spring force $F$.}
    (a) Predictions for select values of $t$. Outputs with $|\psi_2-\psi_1|>4$ (highlighted in red) are extrapolated. 
    (b) Coefficient functions in the equation $\hat{F}(t,\hat{z}_1,\hat{z}_2)=\hat{k}_1(t)\cdot\hat{z}_1-\hat{k}_2(t)\cdot\hat{z}_2$ learned by the SEQL network.
    (c) Latent variable encodings for the force function $F$ learned by (left) the convolutional SEQL network and (right) the ReLU network.
    }
    \label{fig:conv-force}
\end{figure}

\begin{table*}[tbp]
    \centering
    \caption{Learned equations of the SEQL on select $t$ values for the spring force function $F(t,\psi_1,\psi_2)=-\frac{5-t}{2}\cdot(\psi_2-\psi_1)$ in the latent space and transformed to the original parameter space.}
    \begin{tabular}{rrrr}
        \myalign{c}{$t$} & \myalign{c}{True} & \myalign{c}{Learned Latent} & \myalign{c}{Learned Transformed} \\ \hline
        $-2.619$ & $-3.81(\psi_2-\psi_1)$ & $-4.66\hat{z}_1+4.66\hat{z}_2$ & $3.82\hat{\psi}_1-3.82\hat{\psi}_2$ \\
        $-1.095$ & $-3.05(\psi_2-\psi_1)$ & $-3.72\hat{z}_1+3.72\hat{z}_2$ & $3.05\hat{\psi}_1-3.05\hat{\psi}_2$\\
        $0.381$ & $-2.31(\psi_2-\psi_1)$ & $-2.82\hat{z}_1+2.82\hat{z}_2$ & $2.31\hat{\psi}_1-2.31\hat{\psi}_2$\\
        $1.905$ & $-1.55(\psi_2-\psi_1)$ &  $-1.89\hat{z}_1+1.89\hat{z}_2$ & $1.55\hat{\psi}_1-1.55\hat{\psi}_2$ \\
    \end{tabular}
    \label{tab:conv-force-eqs}
\end{table*}

Results for learning the spring force is shown in Figure \ref{fig:conv-force}. All three of the SEQL, HEQL, and ReLU architectures accurately predict the force inside the training domain, but only the SEQL and HEQL networks are able to extrapolate outside of the training regime whereas the ReLU network fails to extrapolate. Additionally, the SEQL network learns the governing equation as shown in Table \ref{tab:conv-force-eqs}, with the learned parametric coefficient plotted in Figure \ref{fig:conv-force}(b) (results for the HEQL are similar). The equations that the SEQL network learns are fairly simple and interpretable, and can be written as  $\hat{F}=\hat{k}_1(t)\hat{z}_1-\hat{k}_2(t)\hat{z}_2$. Upon inspection, we see that $\hat{k}_1(t)\approx \hat{k}_2(t)$ and so we can simplify the learned expression to $\hat{F}=\hat{k}(t)\hat{z}_1-\hat{k}(t)\hat{z}_2$. Thus, the SEQL network has discovered the true force equation underlying the system.

Additionally, while the SEQL network discovers an equation in terms of $\hat{z}_{1,2}$, it also learns a linear mapping of the latent variable to the true position as shown in Figure \ref{fig:conv-force}(c). While there is no explicit constraint or regularization placed on the latent space, because the EQL network must learn to use the latent variable to form the equation, the end-to-end training of the architecture forces the mapping to be an analytical transformation of the original variable, which in this case is a linear mapping. In contrast, whilte it is one-to-one, the latent variable mapping for the ReLU network is not linear since there is no bias to make the mapping linear. Using this linear mapping, we can perform a linear regression to find the approximate relationship between $\hat{z}$ and $\hat{\psi}$ and reconstruct the discovered equation in terms of $\hat{\psi}$, which is shown in the right-most column of Table \ref{tab:conv-force-eqs}.

\begin{figure}
    \centering
    \includegraphics[width=0.8\columnwidth]{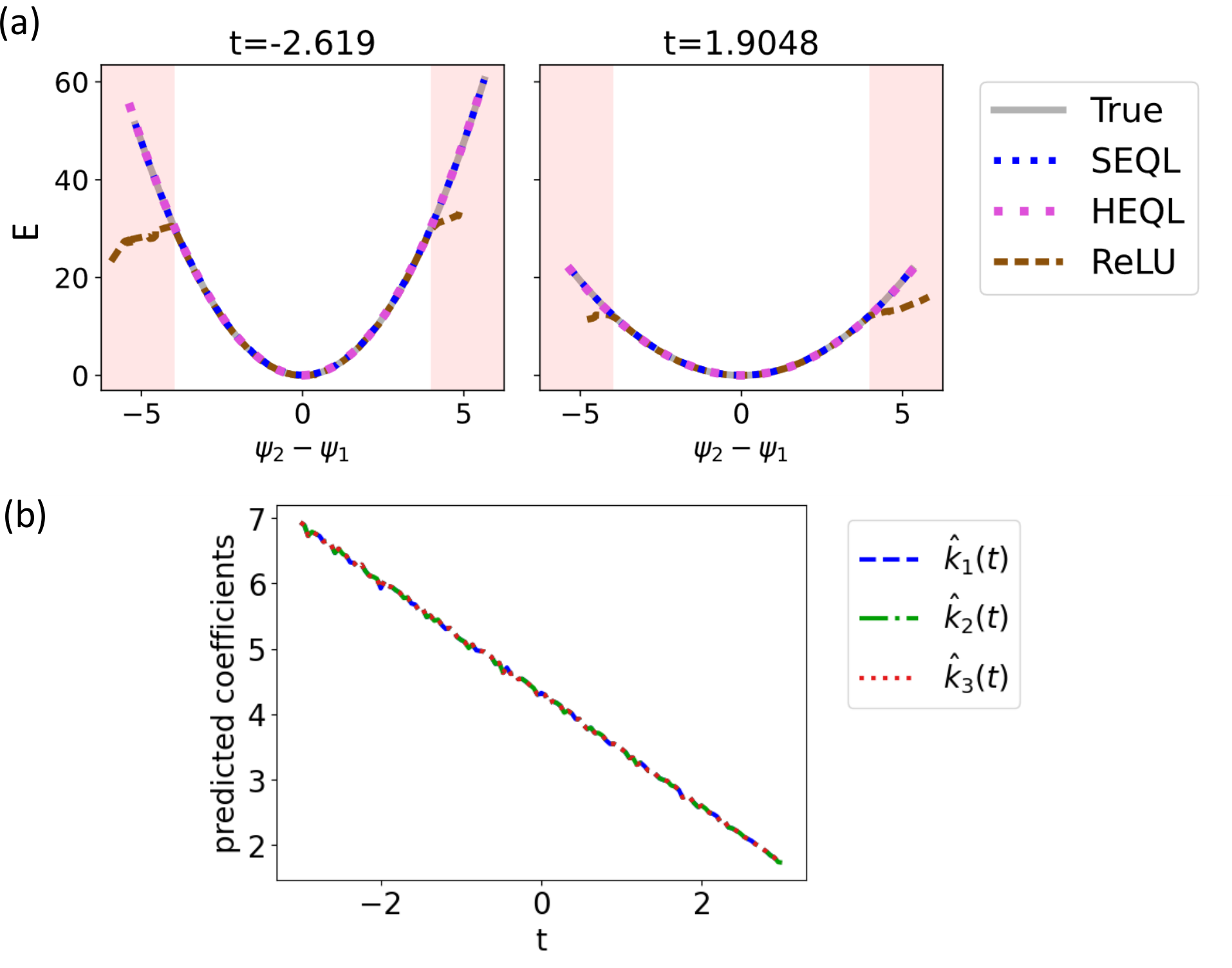}
    \caption{\textbf{Results for learning the spring energy $E$.}
    (a) Predictions for select values of $t$. Outputs with $|\psi_2-\psi_1|>4$ (highlighted in red) are extrapolated. 
    (b) Coefficient functions in the equation $f(t,\hat{z}_1,\hat{z}_2)=\hat{k}_1(t)\cdot\hat{z}_1^2+\hat{k}_2(t)\cdot\hat{z}_2^2-2\hat{k}_3(t)\cdot\hat{z}_1\hat{z}_2$ learned by the HEQL network.}
    \label{fig:conv-potential}
\end{figure}

\begin{table*}[tbp]
    \centering
    \caption{Learned equations of the HEQL network on select $t$ values for the function $E(t,\psi_1,\psi_2)=\frac{5-t}{4}\cdot(\psi_2-\psi_1)^2$ in the latent space and transformed to the original parameter space.}
    \begin{tabular}{rrrr}
        \myalign{c}{$t$} & \myalign{c}{True} & \myalign{c}{Learned Latent} & \myalign{c}{Learned Transformed} \\ \hline
        $-2.619$ & $-1.90(\psi_2-\psi_1)^2$ & $6.59\hat{z}_1^2+6.59\hat{z}_2^2-13.18\hat{z}_1\hat{z}_2+0.02$ & $1.91\hat{\psi}_1^2+1.91\hat{\psi}_2^2-3.82\hat{\psi}_1\hat{\psi}_2+0.02$\\
        $-1.095$ & $-1.52(\psi_2-\psi_1)^2$ & $5.27\hat{z}_1^2+5.27\hat{z}_2^2-10.55\hat{z}_1\hat{z}_2+0.01$ & $1.53\hat{\psi}_1^2+1.53\hat{\psi}_2^2-3.06\hat{\psi}_1\hat{\psi}_2+0.01$ \\
        $0.381$ & $-1.16(\psi_2-\psi_1)^2$ & $4.01\hat{z}_1^2+4.01\hat{z}_2^2-8.02\hat{z}_1\hat{z}_2+0.01$ & $1.16\hat{\psi}_1^2+1.16\hat{\psi}_2^2-2.33\hat{\psi}_1\hat{\psi}_2+0.01$\\
        $1.905$ & $-0.77(\psi_2-\psi_1)^2$ & $2.64\hat{z}_1^2+2.64\hat{z}_2^2-5.28\hat{z}_1\hat{z}_2+0.01$ & $0.77\hat{\psi}_1^2+0.77\hat{\psi}_2^2-1.53\hat{\psi}_1\hat{\psi}_2+0.01$\\
    \end{tabular}
    \label{tab:conv-potential-eqs}
\end{table*}

We see similar results for the spring potential data, this time using the HEQL network, in Figures \ref{fig:conv-potential} and Table \ref{tab:conv-potential-eqs}. Again, the SEQL and HEQL arhictectures are able to extrapolate outside of the training regime whereas the ReLU network fails to extrapolate. Note that in this case, the HEQL learns the equation $\hat{E}(t,\hat{z}_1,\hat{z}_2)=\hat{k}_1(t)\hat{z}_1^2+\hat{k}_2(t)\hat{z}_2^2-2\hat{k}_3(t)\hat{z}_1\hat{z}_2+\epsilon(t)$ where  $\hat{k}_1\approx\hat{k}_2\approx\hat{k}_3$ and $\epsilon$ is small. Thus, the HEQL network has discovered the correct equation.

\section{Discussion}

All results for both architectures can be found in the Appendix. 
Comparing the two architectures, for a moderate number of time steps (e.g. $N_t < 512$) the SEQL has fewer parameters than the HEQL; despite this, however, the HEQL trains on each minibatch $3.7\times$ faster than the SEQL on the analytic equations for our settings of hyper-parameters and network sizes. 
This is likely because the limiting factor is the computation of the activation functions, which must be processed separately for each component of the layer output $h$ (whereas in a conventional neural network the use of a single activation function is able to take advantage of vectorization optimizations). 
For a larger number of time steps, (e.g. $N_t > 512$), the HEQL is more memory-efficient as well since the SEQL parameters scale linearly with the number of time steps. 
Thus, the HEQL is able to scale to larger datasets. 
Future work can include reducing the memory requirements of the MWU inside the HEQL, perhaps by parameterizing the EQL network using lower-rank matrices so that the dimensionality of the MWU output can be reduced.

In terms of the data format, prior methods rely on gridded data \cite{rudy2019,xu2021deep} while both the SEQL and the HEQL allow a variable grid along the varying dimension. 
The HEQL architecture takes this flexibility a step further in that it is able to interpolate in time and make predictions at arbitrary time points, whereas the stacked architecture is fixed to certain time points. 
On the other hand, we find that the stacked architecture is less sensitive to the random initialization and converges more quickly to the solution. 
Thus, the two architectures trade off between performance and flexibility.
One possible direction for future work to bridge this gap is to introduce different learning rate schedules for the EQL network and the MWU in the HEQL architecture, as the EQL network typically requires large learning rates to escape local minima and converge, whereas large learning rates may be detrimental to the MWU.

As mentioned in Section \ref{sec:results}, the SEQL and HEQL architectures are also more flexible than previous approaches in the types of equations that can be discovered.
For example, the previous approaches rely on variants of linear regression, and are thus not able to discover varying coefficients that are inside other functions such as $\sin(f(t)x)$.
Additionally, our approach is able to automatically discover cross terms whereas the SINDy framework relies on these terms being precomputed.

\section{Conclusion}

We have proposed two different variants of the EQL network---the stacked EQL architecture (SEQL) and the hyper EQL architecture (HEQL)---to enable neural network-based symbolic regression of parametric equation where coefficients may vary. We have demonstrated our system on simple analytic equations, PDEs, and a dataset encoded as images, and have found that we are able to discover interpretable equations that can extrapolate outside of the training regime. Our method has the potential to combine the power of deep learning and symbolic regression to enable scientific discovery on complex and high-dimensional datasets.

We note that in our experiments we used simple functions for the varying coefficients for simplicity. However, our method is not constrained to these types of expressions, and the parametric coefficient can more generally be any arbitrary function. Thus, our method can be applied to systems that we know are partially governed by an analytic equation, but partially governed by some other mechanism that may be too complex or noisy to capture. This is similar in spirit to methods for solving PDEs that replace part of the equation with a neural network, often to correct for discretization errors \cite{pathak2020using,kochkov2021machine}. 

The HEQL architecture can be viewed as implementing functional regularization. Functional regularization, which imposes regularization on the learned function rather than on the parameters, is attractive as it is much more intuitive and can lead to more natural methods for tasks such as continual learning \cite{benjamin2018measuring,pan2020continual}. 
It has been explored in neural networks through regularizing the predictions on batches of data \cite{benjamin2018measuring} and through defining the prior over functions rather than weights in the case of Bayesian neural networks \cite{sun2019functional,rudner2020rethinking}. In the case of the EQL network, the coefficients of the resulting equation are typically very simple functions (oftentimes the identity function) of the weights themselves. This means that in practice, the $L_2$ smoothing regularization in the stacked EQL network architecture often implicitly applies to the \textit{function space}, even though we are explicitly applying the regularization in the \textit{weight space}. In the case of the HEQL architecture, the output of fully-connected neural networks will tend to be smooth due to modern training methods such as stochastic gradient descent (which is a topic of great interest in itself), and so the MWU itself acts as a regularization on the function space of the EQL network. 
Given this inherent regularization, another interesting direction for future work would be to characterize the data efficiency of our proposed architectures, especially for sparse datasets.

While the proposed architectures in this work aim to address the challenge of discovering parametric equations, they still share some of the limitations of the original EQL network proposed by \cite{kim2020integration} including sensitivity to random initializations and difficulties with converging when using non-conventional activation functions. The skip connections and $L_0$ regularization in this work improve the convergence behavior compared to the original EQL network, but there is still further room for improvement. For example, the Snake function, defined as $x+\frac{1}{a}\sin^2(ax)$ where $a$ is a learnable parameter, could be used to learn periodic functions while maintaining monotonicity and thus improve convergence \cite{ziyin2020neural}. Pad\'e Activation Units (PAU) \cite{molina2019pad} or the neural arithmetic logic unit (NALU) \cite{Trask2018,schlor2020inalu} could be used to learn rational functions, since the discontinuity of the division operator makes it difficult to straightforwardly incorporate as an activation function. These limitations have not hampered most of the existing proposed models for discovering differential equations, as known differential equations rarely include such terms. However, since such functions are widely prevalent in science and engineering equations, a future direction should explore a more robust way to learn these types of functions.

\section*{Acknowledgments}
We would like to thank Rumen Dangovski, Anka Hu, and Amber Li for insightful discussions and work on related projects.
This work is supported in part by the MIT UROP program, the National Science Foundation under Cooperative Agreement PHY-2019786 (The NSF AI Institute for Artificial Intelligence and Fundamental Interactions, \url{http://iaifi.org/}), the National Defense Science \& Engineering Graduate Fellowship (NDSEG) Program, and the Air Force Office of Scientific Research under the award number FA9550-21-1-0317.
Research was sponsored by the United States Air Force Research Laboratory and the United States Air Force Artificial Intelligence Accelerator and was accomplished under Cooperative Agreement Number FA8750-19-2-1000. The views and conclusions contained in this document are those of the authors and should not be interpreted as representing the official policies, either expressed or implied, of the United States Air Force or the U.S. Government. The U.S. Government is authorized to reproduce and distribute reprints for Government purposes notwithstanding any copyright notation herein.

{\appendices

\section{Architecture Details} \label{app:training}

Each of the SEQL and HEQL consists of 2 hidden layers. The activation functions in each hidden layer consist of:
\begin{equation*} 
[1 (\times 2), g (\times 4), g^2 (\times 4), \sin (2\pi g) (\times 2), g_1*g_2 (\times 2)] 
\end{equation*}
\noindent where the $(\times i)$ indicates the number of times each activation function is duplicated. The $\sin$ function has a multiplier inside so that the functions more accurately represent their respective shapes inside the input domain of $x\in[-1,1]$. The exact number of duplications is arbitrary and does not have a significant impact on the system's performance.

For the HEQL, the MWU consists of a fully-connected neural network with 3 hidden layers of 64, 64, and 256 hidden units, respectively. The hidden layers in the MWU use the ReLU function as the activation.

The network is trained using the RMSProp optimizer and a sum of the MSE loss and regularization. For the HEQL architecture, the regularization is simply the $L_0$ regularization, whereas the SEQL has an additional regularization across time steps to induce smooth functions as described in Section \ref{sec:stack-architecture}

For both learning rate and regularization weight schedules, we use a one cycle policy, as shown in Figure \ref{fig:schedules}. We start off with a small learning rate and regularization to ensure the EQL network settles into a stable configuration containing many different terms such that the network weights do not explode. The learning rate is ramped up to allow the EQL network escape local minima in search of global minima, and the regularization is likewise increased to pare down the number of terms. Finally, we expect the EQL network to have learned the correct equation structure partway through training, and so we decrease learning rate and regularization to fine-tune the weights and optimize primarily for MSE.

To extract the learned equation from the trained EQL network, we can simply multiply the weights by the primitive functions using symbolic mathematics. We implement this using SymPy, which can automatically simplify the expression \cite{sympy}. Additionally, we use a threshholding procedure in the final expression where we drop terms where the coefficient is smaller than a threshhold, which we set to $0.01$.

\begin{figure}
    \centering
    \includegraphics[width=0.45\columnwidth]{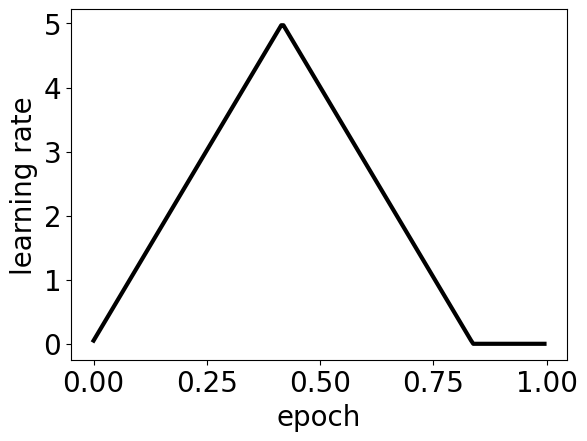}
    \includegraphics[width=0.45\columnwidth]{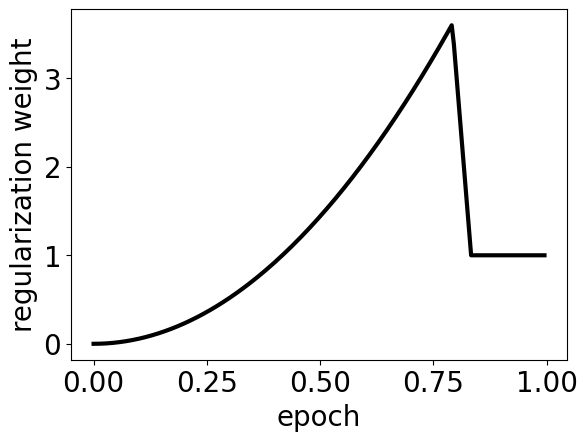}
    \caption{(Left) Learning rate and (right) regularization weight schedules during training relative to \texttt{base\_lr} and \texttt{base\_rw}.}
    \label{fig:schedules}
\end{figure}

All experiments were run on an NVIDIA GeForce RTX 2080 Ti graphics card. On average, the SEQL took 1490 seconds while the HEQL took 413 seconds to perform 34800 mini-batch gradient steps on the analytic tasks for our settings of the architecture and training details. As mentioned in the Discussion section, the HEQL is able to scale more efficiently than the SEQL to larger number of time steps, and thus, larger datasets, since the SEQL scales linearly with the number of time steps. 

\section{Additional Results}

\subsection*{Analytic Expression} \label{app:results-analytic}

For all tests, $512$ training data points with $x\in[-3,3]$ are randomly sampled for each of $128$ fixed, equally-spaced, values of $t\in[-3,3]$ for a total of $512\cdot128=65\,536$ training examples. To test generalization, the parametric EQL architectures are evaluated on $256$ test data points with $x\in[-5,5]$ across the same $128$ fixed values of $t$.

Due to sensitivity of the parametric EQL architectures to the random initialization of network weights, $40$ trials were run for each function. In practice, the networks only need to learn the correct equation once over a reasonable number of trials, since it is possible to construct a validation method that selects the best equation from a set of learned equations. For all the results in this paper, we simply select the trial with the lowest generalization error. Other considerations that can be integrated in the validation process are equation simplicity and prior beliefs about the equation form, for example.

\begin{table*}
    \centering
    \caption{Learned equations for additional analytic expressions.}
    \begin{tabular}{crrrr}
        & & & \multicolumn{2}{c}{Learned equations} \\
        \cmidrule(lr){4-5}
        Benchmark & \myalign{c}{$t$} & \myalign{c}{True} & \myalign{c}{SEQL} & \myalign{c}{HEQL} \\ \hline
        \multirow{4}{*}{$f_3=t\cdot x$} 
            & $-2.619$ & $-2.62x$ & $-2.62x$ & $-2.62x-0.02$\\
            & $-1.095$ & $-1.10x$ &  $-1.10x$ & $-1.10x\hphantom{+00.00}$\\
            & $0.381$ & $0.38x$ & $0.38x$ & $0.38x\hphantom{+00.00}$\\
            & $1.905$ & $1.90x$ & $1.91x$  & $1.91x\hphantom{+00.00}$ \\ \hline
        \multirow{4}{*}{$f_4=t\cdot x^2+3\sin(t)\cdot x$} 
            & $-2.619$ & $-2.62x^2-1.50x$ & $-2.62x^2-1.51x-0.08$ & $-2.62x^2-1.50x+0.01$ \\
            & $-1.095$ & $-1.10x^2-2.67x$ & $-1.09x^2-2.68x-0.08$ & $-1.10x^2-2.66x+0.02$\\
            & $0.381$ & $0.38x^2+1.12x$ & $0.38x^2+1.12x+0.01$ & $0.38x^2+1.12x\hphantom{+00.00}$\\
            & $1.905$ & $1.90x^2+2.83x$ & $1.90x^2+2.85x+0.06$ & $1.90x^2+2.83x\hphantom{+00.00}$ \\ \hline
        \multirow{4}{*}{$f_5=\sin\left(\frac{5+t}{2}\cdot x\right)$} 
            & $-2.619$ & $\sin(1.19x)$ & $\sin(1.19x)$ & $\sin(1.19x)$\\
            & $-1.095$ & $\sin(1.95x)$ & $\sin(1.95x)$ & $\sin(1.95x)$\\
            & $0.381$ & $\sin(2.69x)$ & $\sin(2.69x)$ & $\sin(2.69x)$\\
            & $1.905$ & $\sin(3.45x)$ & $\sin(3.45x)$ & $\sin(3.45x)$\\
    \end{tabular}
    \label{tab:contr-id-eqs}
\end{table*}

\begin{table*}
    \centering
    \caption{Results for analytic expression benchmarks.}
    \begin{tabular}{ccccc}
        & \multicolumn{2}{c}{Test MSE of the best trial} & \multicolumn{2}{c}{Mean (Standard Deviation) Test MSE over all trials}\\
        \cmidrule(lr){2-3} \cmidrule(lr){4-5}
        Benchmark & SEQL & HEQL & SEQL & HEQL\\ \hline
        $f_1$ & $\mathbf{6.97\times10^{-6}}$ & $2.04\times10^{-5}$ & 
            $\mathbf{1.84\times10^{-5}\; (1.97\times10^{-5})}$ & $2.24\times10^{-5}\; (2.53\times10^{-6})$\\
        $f_2$ & $8.98\times10^{-7}$ & $\mathbf{1.22\times10^{-7}}$ & 
            $1.32\times10^{-4}\; (4.14\times10^{-4})$ & $\mathbf{4.82\times10^{-8}\; (1.26\times10^{-8})}$\\
        $f_3$ & $\mathbf{5.28\times10^{-15}}$ & $4.43\times10^{-6}$ & 
            $\mathbf{5.72\times10^{-15}\; (3.83\times10^{-16})}$ & $6.60\times10^{-6}\; (1.45\times10^{-6}$ \\
        $f_4$ &  $\mathbf{9.47\times10^{-6}}$ & $3.17\times10^{-5}$ & 
            $\mathbf{8.32\times10^{-5}\; (4.74\times10^{-4})}$ & $1.04\times10^{-3}\; (6.04\times10^{-3})$\\
        $f_5$ & $1.21\times10^{-7}$ & $\mathbf{3.63\times10^{-8}}$ &
            $2.53\times10^{-5}\; (1.51\times10^{-4})$ & $\mathbf{4.87\times10^{-8}\; (2.29\times10^{-8})}$\\
    \end{tabular}
    \label{tab:analytic-benchmark-mse}
\end{table*}

Additional analytical expression benchmarks and the discovered equations for $f_3$, $f_4$, and $f_5$ by the SEQL and HEQL architectures are listed in Table \ref{tab:contr-id-eqs}.
Both the SEQL and HEQL match the true equations very closely.
We also list various quantitative metrics for these benchmarks in Table \ref{tab:analytic-benchmark-mse}, including the MSE on the training and test datasets, as well as the mean and standard deviation of the test MSE over all the trials.
The aggregate metrics over all the trials tend to be similar in magnitude to the metric of the best trial for many of the benchmarks, which signifies that the model has learned the correct equation is a large majority of the trials.
When a model fails to learn the correct equation, the MSE on the test dataset tends to be several orders of magnitude larger than that of the best trial, which would skew the mean and standard deviation of the MSE.
Interestingly, there is no clear trend on whether the SEQL or the HEQL performs better.

For simple benchmarks such as $f_3=tx$, both the SEQL and HEQL architectures are able to find the correct equation structure nearly 100\% of the time, even if the accuracy of the varying coefficients may vary slightly. However, in other cases such as $f_4$, the HEQL will sometimes learn the equation:
\begin{equation*}
    \hat{f}_{4,HEQL}=a(t)x^2+b(t)x+c(t)\sin(d(t)x+e(t))
\end{equation*}
where $d(t)$ is small. This is likely because the architecture is using the approximately linear region of the low-frequency sinusoid, and adding it to the $b(t)x$ term. We also note that the SEQL is able to find the correct equation more often. Another interesting failure mode is in the case of the sinusoid functions (i.e., $f_2$ and $f_5$) where the HEQL will somtimes learn the equation:
\begin{equation*}
    \hat{f}_{5, HEQL}=a(t)\sin(b(t)x)+c(t)\sin(d(t)x)
\end{equation*}
where $b(t)\approx d(t)$ and $a(t) + c(t) \approx 1$. The symbolic manipulation is unable to combine the two terms, but one can see upon inspection that the HEQL has learned the correct form of the varying parameters.

\subsection*{Partial Differential Equations (PDEs)}

For the advection-diffusion equation, data was sampled from $256$ different points in the $x$-domain and $512$ different points in the $t$-domain, for a total of $256\cdot512=131\,072$ examples. The equation is solved numerically using a spectral method on the domain $x\in[-5,5]$ and $t\in[0,5]$ with $f(x)=-1.5+\cos\left(\frac{2\pi x}{5}\right)$ and $\epsilon=0.1$ using code from \cite{rudy2019}. 

For the Burgers' equation, data was sampled from $512$ different points in the $x$-domain and $256$ different points in the $t$-domain, for a total of $512\cdot256=131\,072$ examples. The equation was solved numerically using a spectral method on the domain $x\in[-8,8]$ and $t\in[0,10]$ using code from \cite{rudy2019}. Similar to the analytic expression experiments, $80$ trials were run for each equation and the trial with the lowest training error was selected.

\bibliographystyle{IEEEtranN}
\bibliography{sym_sources}

\begin{thebibliography}{45}
\providecommand{\natexlab}[1]{#1}
\providecommand{\url}[1]{#1}
\csname url@samestyle\endcsname
\providecommand{\newblock}{\relax}
\providecommand{\bibinfo}[2]{#2}
\providecommand{\BIBentrySTDinterwordspacing}{\spaceskip=0pt\relax}
\providecommand{\BIBentryALTinterwordstretchfactor}{4}
\providecommand{\BIBentryALTinterwordspacing}{\spaceskip=\fontdimen2\font plus
\BIBentryALTinterwordstretchfactor\fontdimen3\font minus
  \fontdimen4\font\relax}
\providecommand{\BIBforeignlanguage}[2]{{%
\expandafter\ifx\csname l@#1\endcsname\relax
\typeout{** WARNING: IEEEtranN.bst: No hyphenation pattern has been}%
\typeout{** loaded for the language `#1'. Using the pattern for}%
\typeout{** the default language instead.}%
\else
\language=\csname l@#1\endcsname
\fi
#2}}
\providecommand{\BIBdecl}{\relax}
\BIBdecl

\bibitem[Griffiths(2013)]{Griffiths:1492149}
\BIBentryALTinterwordspacing
D.~J. Griffiths, \emph{{Introduction to electrodynamics; 4th ed.}}\hskip 1em
  plus 0.5em minus 0.4em\relax Boston, MA: Pearson, 2013, re-published by
  Cambridge University Press in 2017. [Online]. Available:
  \url{https://cds.cern.ch/record/1492149}
\BIBentrySTDinterwordspacing

\bibitem[Schmidt and Lipson(2009)]{Schmidt2009}
\BIBentryALTinterwordspacing
M.~Schmidt and H.~Lipson, ``{Distilling free-form natural laws from
  experimental data.}'' \emph{Science (New York, N.Y.)}, vol. 324, no. 5923,
  pp. 81--5, apr 2009. [Online]. Available:
  \url{http://www.ncbi.nlm.nih.gov/pubmed/19342586}
\BIBentrySTDinterwordspacing

\bibitem[Koza(1994)]{Koza1994}
\BIBentryALTinterwordspacing
J.~Koza, ``{Genetic programming as a means for programming computers by natural
  selection},'' \emph{Statistics and Computing}, vol.~4, no.~2, pp. 87--112,
  jun 1994. [Online]. Available:
  \url{http://link.springer.com/10.1007/BF00175355}
\BIBentrySTDinterwordspacing

\bibitem[Udrescu and Tegmark(2020)]{udrescu2020ai}
S.-M. Udrescu and M.~Tegmark, ``Ai feynman: A physics-inspired method for
  symbolic regression,'' \emph{Science Advances}, vol.~6, no.~16, p. eaay2631,
  2020.

\bibitem[Brunton et~al.(2016)Brunton, Proctor, and
  Kutz]{brunton2016discovering}
S.~L. Brunton, J.~L. Proctor, and J.~N. Kutz, ``Discovering governing equations
  from data by sparse identification of nonlinear dynamical systems,''
  \emph{Proceedings of the national academy of sciences}, vol. 113, no.~15, pp.
  3932--3937, 2016.

\bibitem[Champion et~al.(2019)Champion, Lusch, Kutz, and
  Brunton]{champion2019data}
K.~Champion, B.~Lusch, J.~N. Kutz, and S.~L. Brunton, ``Data-driven discovery
  of coordinates and governing equations,'' \emph{Proceedings of the National
  Academy of Sciences}, vol. 116, no.~45, pp. 22\,445--22\,451, 2019.

\bibitem[Long et~al.(2019)Long, Lu, and Dong]{long2019pde}
Z.~Long, Y.~Lu, and B.~Dong, ``Pde-net 2.0: Learning pdes from data with a
  numeric-symbolic hybrid deep network,'' \emph{Journal of Computational
  Physics}, vol. 399, p. 108925, 2019.

\bibitem[Lu et~al.(2021)Lu, Ari{\~n}o, and
  Solja{\v{c}}i{\'c}]{lu2021discovering}
P.~Y. Lu, J.~Ari{\~n}o, and M.~Solja{\v{c}}i{\'c}, ``Discovering sparse
  interpretable dynamics from partial observations,'' \emph{arXiv preprint
  arXiv:2107.10879}, 2021.

\bibitem[Cranmer et~al.(2020)Cranmer, Sanchez~Gonzalez, Battaglia, Xu, Cranmer,
  Spergel, and Ho]{cranmer2020discovering}
M.~Cranmer, A.~Sanchez~Gonzalez, P.~Battaglia, R.~Xu, K.~Cranmer, D.~Spergel,
  and S.~Ho, ``Discovering symbolic models from deep learning with inductive
  biases,'' \emph{Advances in Neural Information Processing Systems}, vol.~33,
  pp. 17\,429--17\,442, 2020.

\bibitem[Martius and Lampert(2016)]{Martius2016}
\BIBentryALTinterwordspacing
G.~Martius and C.~H. Lampert, ``{Extrapolation and learning equations},''
  \emph{arXiv preprint arXiv:1610.02995}, oct 2016. [Online]. Available:
  \url{http://arxiv.org/abs/1610.02995}
\BIBentrySTDinterwordspacing

\bibitem[Sahoo et~al.(2018)Sahoo, Lampert, and Martius]{sahoo2018learning}
S.~Sahoo, C.~Lampert, and G.~Martius, ``Learning equations for extrapolation
  and control,'' in \emph{International Conference on Machine Learning}.\hskip
  1em plus 0.5em minus 0.4em\relax PMLR, 2018, pp. 4442--4450.

\bibitem[Kim et~al.(2020)Kim, Lu, Mukherjee, Gilbert, Jing, {\v{C}}eperi{\'c},
  and Solja{\v{c}}i{\'c}]{kim2020integration}
S.~Kim, P.~Y. Lu, S.~Mukherjee, M.~Gilbert, L.~Jing, V.~{\v{C}}eperi{\'c}, and
  M.~Solja{\v{c}}i{\'c}, ``Integration of neural network-based symbolic
  regression in deep learning for scientific discovery,'' \emph{IEEE
  Transactions on Neural Networks and Learning Systems}, vol.~32, no.~9, pp.
  4166--4177, 2020.

\bibitem[Costa et~al.(2020)Costa, Dangovski, Dugan, Kim, Goyal,
  Solja{\v{c}}i{\'c}, and Jacobson]{costa2020fast}
A.~Costa, R.~Dangovski, O.~Dugan, S.~Kim, P.~Goyal, M.~Solja{\v{c}}i{\'c}, and
  J.~Jacobson, ``Fast neural models for symbolic regression at scale,''
  \emph{arXiv preprint arXiv:2007.10784}, 2020.

\bibitem[Griffiths and Schroeter(2018)]{griffiths_introduction_2018}
D.~J. Griffiths and D.~F. Schroeter, \emph{Introduction to quantum mechanics},
  third edition~ed.\hskip 1em plus 0.5em minus 0.4em\relax Cambridge ; New
  York, NY: Cambridge University Press, 2018.

\bibitem[Zhang et~al.(2002)Zhang, Jiang, and Wang]{zhang2002recurrent}
Y.~Zhang, D.~Jiang, and J.~Wang, ``A recurrent neural network for solving
  sylvester equation with time-varying coefficients,'' \emph{IEEE Transactions
  on Neural Networks}, vol.~13, no.~5, pp. 1053--1063, 2002.

\bibitem[Yan and Konotop(2009)]{yan2009exact}
Z.~Yan and V.~Konotop, ``Exact solutions to three-dimensional generalized
  nonlinear schr{\"o}dinger equations with varying potential and
  nonlinearities,'' \emph{Physical Review E}, vol.~80, no.~3, p. 036607, 2009.

\bibitem[Rudy et~al.(2019)Rudy, Alla, Brunton, and Kutz]{rudy2019}
\BIBentryALTinterwordspacing
S.~Rudy, A.~Alla, S.~L. Brunton, and J.~N. Kutz, ``Data-driven identification
  of parametric partial differential equations,'' \emph{SIAM Journal on Applied
  Dynamical Systems}, vol.~18, no.~2, pp. 643--660, 2019. [Online]. Available:
  \url{https://doi.org/10.1137/18M1191944}
\BIBentrySTDinterwordspacing

\bibitem[Xu et~al.(2021)Xu, Zhang, and Zeng]{xu2021deep}
H.~Xu, D.~Zhang, and J.~Zeng, ``Deep-learning of parametric partial
  differential equations from sparse and noisy data,'' \emph{Physics of
  Fluids}, vol.~33, no.~3, p. 037132, 2021.

\bibitem[Luo et~al.(2021)Luo, Liu, Chen, Hu, and Zhu]{luo2021ko}
Y.~Luo, Q.~Liu, Y.~Chen, W.~Hu, and J.~Zhu, ``Ko-pde: Kernel optimized
  discovery of partial differential equations with varying coefficients,''
  \emph{arXiv preprint arXiv:2106.01078}, 2021.

\bibitem[Louizos et~al.(2017)Louizos, Welling, and Kingma]{Louizos2017}
\BIBentryALTinterwordspacing
C.~Louizos, M.~Welling, and D.~P. Kingma, ``{Learning Sparse Neural Networks
  through {\$}L{\_}0{\$} Regularization},'' \emph{arXiv preprint
  arXiv:1712.01312}, dec 2017. [Online]. Available:
  \url{https://arxiv.org/abs/1712.01312}
\BIBentrySTDinterwordspacing

\bibitem[Maddison et~al.(2016)Maddison, Mnih, and Teh]{maddison2016concrete}
C.~J. Maddison, A.~Mnih, and Y.~W. Teh, ``The concrete distribution: A
  continuous relaxation of discrete random variables,'' \emph{arXiv preprint
  arXiv:1611.00712}, 2016.

\bibitem[He et~al.(2016)He, Zhang, Ren, and Sun]{he2016deep}
K.~He, X.~Zhang, S.~Ren, and J.~Sun, ``Deep residual learning for image
  recognition,'' in \emph{Proceedings of the IEEE conference on computer vision
  and pattern recognition}, 2016, pp. 770--778.

\bibitem[Huang et~al.(2017)Huang, Liu, Van Der~Maaten, and
  Weinberger]{huang2017densely}
G.~Huang, Z.~Liu, L.~Van Der~Maaten, and K.~Q. Weinberger, ``Densely connected
  convolutional networks,'' in \emph{Proceedings of the IEEE conference on
  computer vision and pattern recognition}, 2017, pp. 4700--4708.

\bibitem[Ha et~al.(2016)Ha, Dai, and Le]{ha2016hypernetworks}
D.~Ha, A.~Dai, and Q.~V. Le, ``Hypernetworks,'' \emph{arXiv preprint
  arXiv:1609.09106}, 2016.

\bibitem[Andrychowicz et~al.(2016)Andrychowicz, Denil, Gomez, Hoffman, Pfau,
  Schaul, Shillingford, and De~Freitas]{andrychowicz2016learning}
M.~Andrychowicz, M.~Denil, S.~Gomez, M.~W. Hoffman, D.~Pfau, T.~Schaul,
  B.~Shillingford, and N.~De~Freitas, ``Learning to learn by gradient descent
  by gradient descent,'' \emph{Advances in neural information processing
  systems}, vol.~29, 2016.

\bibitem[Munkhdalai and Yu(2017)]{munkhdalai2017meta}
T.~Munkhdalai and H.~Yu, ``Meta networks,'' in \emph{International Conference
  on Machine Learning}.\hskip 1em plus 0.5em minus 0.4em\relax PMLR, 2017, pp.
  2554--2563.

\bibitem[Ravi and Larochelle(2017)]{ravi2017optimization}
\BIBentryALTinterwordspacing
S.~Ravi and H.~Larochelle, ``Optimization as a model for few-shot learning,''
  in \emph{International Conference on Learning Representations}, 2017.
  [Online]. Available: \url{https://openreview.net/forum?id=rJY0-Kcll}
\BIBentrySTDinterwordspacing

\bibitem[Hospedales et~al.(2020)Hospedales, Antoniou, Micaelli, and
  Storkey]{hospedales2020meta}
T.~Hospedales, A.~Antoniou, P.~Micaelli, and A.~Storkey, ``Meta-learning in
  neural networks: A survey,'' \emph{arXiv preprint arXiv:2004.05439}, 2020.

\bibitem[Chen et~al.(2018)Chen, Rubanova, Bettencourt, and
  Duvenaud]{chen2018neural}
R.~T. Chen, Y.~Rubanova, J.~Bettencourt, and D.~K. Duvenaud, ``Neural ordinary
  differential equations,'' \emph{Advances in neural information processing
  systems}, vol.~31, 2018.

\bibitem[de~Avila Belbute-Peres et~al.(2021)de~Avila Belbute-Peres, Chen, and
  Sha]{de2021hyperpinn}
F.~de~Avila Belbute-Peres, Y.-f. Chen, and F.~Sha, ``Hyperpinn: Learning
  parameterized differential equations with physics-informed hypernetworks,''
  in \emph{The Symbiosis of Deep Learning and Differential Equations}, 2021.

\bibitem[Nakkiran et~al.(2021)Nakkiran, Kaplun, Bansal, Yang, Barak, and
  Sutskever]{nakkiran2021deep}
P.~Nakkiran, G.~Kaplun, Y.~Bansal, T.~Yang, B.~Barak, and I.~Sutskever, ``Deep
  double descent: Where bigger models and more data hurt,'' \emph{Journal of
  Statistical Mechanics: Theory and Experiment}, vol. 2021, no.~12, p. 124003,
  2021.

\bibitem[Liu et~al.(2020)Liu, Jiang, Bai, Chen, and Wang]{liu2020understanding}
J.~Liu, G.~Jiang, Y.~Bai, T.~Chen, and H.~Wang, ``Understanding why neural
  networks generalize well through gsnr of parameters,'' \emph{arXiv preprint
  arXiv:2001.07384}, 2020.

\bibitem[Liu et~al.(2022)Liu, Mao, Wu, Feichtenhofer, Darrell, and
  Xie]{liu2022convnet}
Z.~Liu, H.~Mao, C.-Y. Wu, C.~Feichtenhofer, T.~Darrell, and S.~Xie, ``A convnet
  for the 2020s,'' \emph{arXiv preprint arXiv:2201.03545}, 2022.

\bibitem[Jakubovitz et~al.(2019)Jakubovitz, Giryes, and
  Rodrigues]{jakubovitz2019generalization}
D.~Jakubovitz, R.~Giryes, and M.~R. Rodrigues, ``Generalization error in deep
  learning,'' in \emph{Compressed sensing and its applications}.\hskip 1em plus
  0.5em minus 0.4em\relax Springer, 2019, pp. 153--193.

\bibitem[Meurer et~al.(2017)Meurer, Smith, Paprocki, \v{C}ert\'{i}k, Kirpichev,
  Rocklin, Kumar, Ivanov, Moore, Singh, Rathnayake, Vig, Granger, Muller,
  Bonazzi, Gupta, Vats, Johansson, Pedregosa, Curry, Terrel, Rou\v{c}ka, Saboo,
  Fernando, Kulal, Cimrman, and Scopatz]{sympy}
\BIBentryALTinterwordspacing
A.~Meurer, C.~P. Smith, M.~Paprocki, O.~\v{C}ert\'{i}k, S.~B. Kirpichev,
  M.~Rocklin, A.~Kumar, S.~Ivanov, J.~K. Moore, S.~Singh, T.~Rathnayake,
  S.~Vig, B.~E. Granger, R.~P. Muller, F.~Bonazzi, H.~Gupta, S.~Vats,
  F.~Johansson, F.~Pedregosa, M.~J. Curry, A.~R. Terrel, v.~Rou\v{c}ka,
  A.~Saboo, I.~Fernando, S.~Kulal, R.~Cimrman, and A.~Scopatz, ``Sympy:
  symbolic computing in python,'' \emph{PeerJ Computer Science}, vol.~3, p.
  e103, Jan. 2017. [Online]. Available:
  \url{https://doi.org/10.7717/peerj-cs.103}
\BIBentrySTDinterwordspacing

\bibitem[Pathak et~al.(2020)Pathak, Mustafa, Kashinath, Motheau, Kurth, and
  Day]{pathak2020using}
J.~Pathak, M.~Mustafa, K.~Kashinath, E.~Motheau, T.~Kurth, and M.~Day, ``Using
  machine learning to augment coarse-grid computational fluid dynamics
  simulations,'' \emph{arXiv preprint arXiv:2010.00072}, 2020.

\bibitem[Kochkov et~al.(2021)Kochkov, Smith, Alieva, Wang, Brenner, and
  Hoyer]{kochkov2021machine}
D.~Kochkov, J.~A. Smith, A.~Alieva, Q.~Wang, M.~P. Brenner, and S.~Hoyer,
  ``Machine learning--accelerated computational fluid dynamics,''
  \emph{Proceedings of the National Academy of Sciences}, vol. 118, no.~21,
  2021.

\bibitem[Benjamin et~al.(2018)Benjamin, Rolnick, and
  Kording]{benjamin2018measuring}
A.~S. Benjamin, D.~Rolnick, and K.~Kording, ``Measuring and regularizing
  networks in function space,'' \emph{arXiv preprint arXiv:1805.08289}, 2018.

\bibitem[Pan et~al.(2020)Pan, Swaroop, Immer, Eschenhagen, Turner, and
  Khan]{pan2020continual}
P.~Pan, S.~Swaroop, A.~Immer, R.~Eschenhagen, R.~Turner, and M.~E.~E. Khan,
  ``Continual deep learning by functional regularisation of memorable past,''
  \emph{Advances in Neural Information Processing Systems}, vol.~33, pp.
  4453--4464, 2020.

\bibitem[Sun et~al.(2019)Sun, Zhang, Shi, and Grosse]{sun2019functional}
S.~Sun, G.~Zhang, J.~Shi, and R.~Grosse, ``Functional variational bayesian
  neural networks,'' \emph{arXiv preprint arXiv:1903.05779}, 2019.

\bibitem[Rudner et~al.(2020)Rudner, Chen, and Gal]{rudner2020rethinking}
T.~G. Rudner, Z.~Chen, and Y.~Gal, ``Rethinking function-space variational
  inference in bayesian neural networks,'' in \emph{Third Symposium on Advances
  in Approximate Bayesian Inference}, 2020.

\bibitem[Ziyin et~al.(2020)Ziyin, Hartwig, and Ueda]{ziyin2020neural}
L.~Ziyin, T.~Hartwig, and M.~Ueda, ``Neural networks fail to learn periodic
  functions and how to fix it,'' \emph{Advances in Neural Information
  Processing Systems}, vol.~33, pp. 1583--1594, 2020.

\bibitem[Molina et~al.(2019)Molina, Schramowski, and Kersting]{molina2019pad}
A.~Molina, P.~Schramowski, and K.~Kersting, ``Pad$\backslash$'e activation
  units: End-to-end learning of flexible activation functions in deep
  networks,'' \emph{arXiv preprint arXiv:1907.06732}, 2019.

\bibitem[Trask et~al.(2018)Trask, Hill, Reed, Rae, Dyer, and
  Blunsom]{Trask2018}
A.~Trask, F.~Hill, S.~Reed, J.~Rae, C.~Dyer, and P.~Blunsom, ``{Neural
  Arithmetic Logic Units},'' \emph{Advances in neural information processing
  systems}, vol.~31, 2018.

\bibitem[Schl{\"o}r et~al.(2020)Schl{\"o}r, Ring, and Hotho]{schlor2020inalu}
D.~Schl{\"o}r, M.~Ring, and A.~Hotho, ``inalu: Improved neural arithmetic logic
  unit,'' \emph{Frontiers in Artificial Intelligence}, vol.~3, p.~71, 2020.

\end{thebibliography}


 




\vfill

\end{document}